\documentclass{article}

\usepackage[preprint]{corl_2026} %

\usepackage{graphicx}
\usepackage{amsmath}
\usepackage{amssymb}
\usepackage{subcaption}  %
\usepackage{algorithm}
\usepackage{algpseudocode}
\usepackage{tabularx}
\usepackage{booktabs}
\usepackage{wrapfig}
\usepackage{enumitem}
\usepackage{float}
\usepackage{placeins}
\usepackage{url}

\usepackage[capitalise]{cleveref}

\usepackage{enumitem} 

\title{Training and Evaluating Diffusion Policies with \\ Long Context Lengths}

\usepackage{color-edits}
\definecolor{darkgreen}{rgb}{0.0, 0.65, 0.0}
\addauthor{aa}{red}
\addauthor{aw}{blue}
\addauthor{tk}{darkgreen}
\addauthor{mz}{cyan}
\addauthor{cb}{cyan}
\addauthor{akd}{cyan}
\addauthor{pp}{magenta}
\addauthor{ao}{magenta}
\addauthor{rt}{orange}

\renewcommand{\thefootnote}{\fnsymbol{footnote}}

\author{
\begin{tabular}{c}
\\[0.15em]
{\bfseries
Abhinav Agarwal\footnotemark[1] \quad
Adam Wei\footnotemark[2] \quad
Taylan Kargin\footnotemark[2] \quad
Michael Zeng \quad
Cole Becker
}
\\[0.15em]
{\bfseries
Arif Kerem Day\i \quad
Pablo Parrilo \quad
Asuman Ozdaglar \quad
Russ Tedrake
}
\\[0.35em]
{\normalfont\normalsize Massachusetts Institute of Technology}
\\[0.35em]
{\normalfont\ttfamily\small \url{https://dp-with-long-context.github.io/}}
\end{tabular}
}

\begin{document}
\maketitle

\vspace{-1.25em}

\footnotetext[1]{Corresponding author: \texttt{abhi\_ag@mit.edu}. \qquad
\textsuperscript{\dag}Equal contribution.}

\renewcommand{\thefootnote}{\arabic{footnote}}

\begin{abstract}
    Imitation learning has enabled highly-dexterous robotic manipulation from RGB observations. Policies trained with these methods, however, typically condition robot actions on only a short history of observations. These policies cannot solve tasks that require memory and can get stuck repeatedly executing the same failing motions. In this work, we first benchmark policy performance as context length is incrementally increased from short to long, across a spectrum of tasks with varying local stability and memory requirements, and in multiple data regimes. To our knowledge, this is the first study to investigate context length for Diffusion Policies at this level of detail. Our results challenge prior claims: naively scaling context length is not as brittle as advertised in literature. With an appropriate conditioning method and denoising backbone (UNet+Cross-Attention), single-task policies achieve high success rates on many tasks in the usual data regime even with naive scaling. Next, we propose a training algorithm to jointly train policies at multiple context lengths, further reducing the sample complexity of long-context learning. Finally, we apply our findings to re-evaluate some previously proposed solutions to long-context imitation learning. 
\end{abstract}

\section{Introduction}
\label{sec:introduction}

\begin{figure}[htbp]
    \centering
    \includegraphics[width=0.9\linewidth]{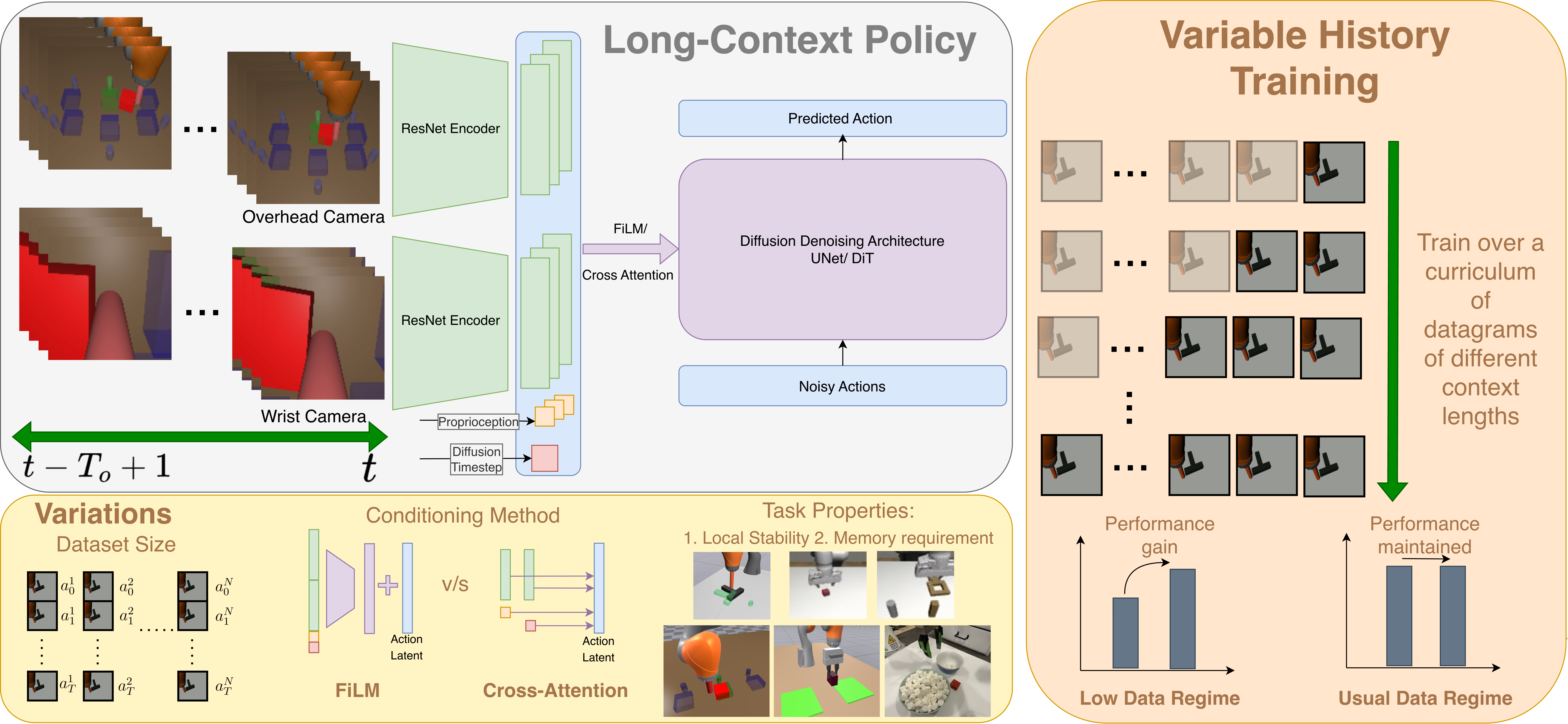}
    \caption{\textbf{Left: }We provide extensive evaluation of key interdependent factors when naively scaling context length for Diffusion Policies \cite{diffusion_policy} by varying data scale, conditioning architecture, and task properties and find that naive scaling doesn't catastrophically fail in many cases, especially with the right architecture. \textbf{Right: }We propose training policies on a curriculum of context lengths as opposed to just one fixed context length, and find that on most tasks we evaluate, this method improves performance in the low data regime while maintaining performance in higher data regimes.}
    \label{fig:anchor_figure}
\end{figure}

Imitation learning has been at the forefront of rapid advancement in generalizable dexterous manipulation of highly complicated, long-horizon, and difficult to model tasks \cite{diffusion_policy, zhao2023learningfinegrainedbimanualmanipulation, trilbmteam2025carefulexaminationlargebehavior, intelligence2025pi05visionlanguageactionmodelopenworld, geminiteam2024geminifamilyhighlycapable}. It is fascinating how these policies perform remarkably well despite conditioning actions on just a truncated history of observations (typically no more than current and a recent past frame). Intuitively, short-context policies can suffer from a number of pitfalls. For instance, such truncated context can make it difficult to recover task-relevant state or infer temporal quantities such as motion phase, and can lead to repeated execution of a failing strategy rather than corrective behavior.

Despite these drawbacks, imitation learning policies with short context lengths are considered the state-of-the-art in robotic manipulation. A major reason for this is the observed difficulty of training long-context policies. Recent work has explored extending policy memory through auxiliary losses \cite{torne2025learninglongcontextdiffusionpolicies}, language-guided retrieval mechanisms \cite{sridhar2025memerscalingmemoryrobot, torne2026memmultiscaleembodiedmemory}, and VLM-based frame selection \cite{mark2026bpplongcontextrobotimitation}. While demonstrating improvement, some of these approaches rely on \emph{heuristic compression and filtering of observation history} prior to passing observations to the policy. A frequently stated reason for such design choices is the brittleness of naively scaling observation history \cite{sridhar2025memerscalingmemoryrobot, torne2025learninglongcontextdiffusionpolicies}. 

We ask whether such a conclusion about naive scaling is premature. While recent works have proposed solutions to long-context learning, \emph{a detailed investigation into key factors impacting long-context policy learning is largely absent from literature}. \citet{mark2026bpplongcontextrobotimitation} provide limited ablations on data scaling and action chunking and agree with us that naive scaling could succeed. However, detailed experiments investigating the interaction between important design choices and establishing how these choices impact naive history scaling remain necessary. 

Motivated by this gap in literature, we ask: \emph{how do the key factors impacting robotic imitation learning such as task properties, data scale, and model architecture impact policy performance as context length is increased from short to long?} We answer this question empirically for state-of-the-art Diffusion Policy \cite{diffusion_policy} method, showing noticeable trends: with the right conditioning architecture and data scale, naively scaling history does not show catastrophic drop in performance. Moreover, scaling data substantially mitigates the long-context learning problem, with the sample complexity needed dictated by difficulty of the manipulation primitive and model architecture used. 

Simultaneously, we recognize the need to provide solutions for cases where task specific data might be limited. Therefore, through our empirical investigation we develop insights into \emph{why long-context policies fail when they do.} Building on that, we present a new algorithm to jointly train policies over a curriculum of context lengths. 

\subsection{Statement of Contributions}
\label{subsec:statement_of_contributions}
In this work, we make the following contributions: 
\begin{enumerate}[leftmargin=1.5em, itemsep=1pt, topsep=0pt, parsep=0pt]
    \item \textbf{A systematic study of interdependent factors in long-context learning: }In \cref{sec:performance_trends}, we show comparative policy evaluation on 5 tasks differing in manipulation stability and context length requirements, at 3 data scales per task, and at multiple context lengths. In \cref{sec:cross_attention_conditioning}, we find that with an appropriate policy architecture (UNet+Cross-Attention) and data scale, naively scaling context length is not too brittle. We show examples of tasks which exhibit consistent or improving performance through simple architectural changes as context length is increased. 
    \item \textbf{A new training recipe for the limited-data regime: }In \cref{sec:variable_history_training}, we propose a new training algorithm to jointly train a policy on a curriculum of context lengths as opposed to one long or short context length. Depending on task, we show 1.25x-2x improvement over strongest naive scaling baseline in the low data regime and maintained performance in other data regimes. 
    \item \textbf{Revisiting Past-Token Prediction \cite{torne2025learninglongcontextdiffusionpolicies}: }In \cref{subsec:revisiting_ptp}, we apply our findings and further investigations to compare our work against ideas presented by \citet{torne2025learninglongcontextdiffusionpolicies} and revisit the underlying mechanism of past-action prediction as an auxiliary loss to help with long-context learning. 
\end{enumerate}

Overall, we present comprehensive takeaways for design choices that strongly impact long-context imitation learning, derived by training and evaluating close to 200 policies across various environments, data scales, context lengths, model architectures, and learning algorithms.

\subsection{Related Works}
\label{sec:related_works}
We review some key related works here, with additional references presented in Appendix~\ref{appx:additional_related_works}.

\textbf{Long Context Lengths for Policies Using Diffusion and Flow: }\citet{torne2025learninglongcontextdiffusionpolicies} argue predicting past actions provides temporal consistency. They freeze the observation encoder to save GPU memory and do not attribute success of their policies to this. In our reproduction of their work, however, we note that freezing the observation encoder is contributing to success (Section \ref{subsec:revisiting_ptp}). More recently, the use of VLMs has been explored for contextual summary/filtering \cite{sridhar2025memerscalingmemoryrobot, mark2026bpplongcontextrobotimitation}. While showing improvements, these works rely on language annotations and assume that the most recent observation is sufficient for learning the fine manipulation skills, an assumption that can be questioned given the use of short (as opposed to just most recent) context lengths in some landmark works \cite{diffusion_policy, trilbmteam2025carefulexaminationlargebehavior}. \citet{dai2026robommebenchmarkingunderstandingmemory} benchmark policy performance on tasks with different memory requirements. While they study memory augmentations built over short-context policies with multi-task pre-training, we focus on design choices relevant to long context lengths in state-of-the-art Diffusion Policies \cite{diffusion_policy}.

\textbf{Long Context Lengths for Other Policies: }Information trace overlay \cite{zheng2025tracevlavisualtraceprompting} and explicit distinction between perceptual and cognitive tokens \cite{shi2026memoryvlaperceptualcognitivememoryvisionlanguageaction} have been explored to expand OpenVLA \cite{kim2024openvlaopensourcevisionlanguageactionmodel} memory. \citet{guhur2022instructiondrivenhistoryawarepoliciesrobotic} suggest the use of expressive transformer models to use full task history but primarily demonstrate success on relatively short horizon tasks. Other prior works have attributed failure of context length longer than one to \emph{causal confusion} \cite{dehaan2019causalconfusionimitationlearning, wen2020fightingcopycatagentsbehavioral}. Rigorous work to establish how well this phenomenon holds for Diffusion Policy remains to be done. 

\section{Preliminaries}
\label{sec:preliminaries}
\subsection{Diffusion Policy}
\label{preliminaries:diffusion_policy}
Diffusion Policies \cite{diffusion_policy}, a state-of-the-art imitation learning method, learn a denoiser to sample past and future robot actions conditioned on a history of past observations to accomplish a desired task 
\begin{equation}
    \mathbb{P}(\mathbf{a}[t\!-T_{p}^{past}\!+\!1:t\!+\!T_p] \mid \mathbf{o}[t\!-\!T_o\!+\!1:t]) \label{eq:diffusion_action_distribution}
\end{equation}
by minimizing the denoising loss in \cite{ho2020denoisingdiffusionprobabilisticmodels}. $t$ is the current timestep, $T_p$ is the prediction horizon, $T_o$ is the context length, $\mathbf{o}$ denotes the observations (often images and proprioception), and $\mathbf{a}$ denotes the actions. \citet{diffusion_policy} suggest predicting a chunk comprising of all past actions and the desired future actions ($T^{past}_p = T_o$). Further details about diffusion model training are in Appendix~\ref{appx:diffusion_policy_details}. Our objective in this work is to\textbf{ a)} comment on the impact of key factors as $T_o$ is increased from short to long and \textbf{b)} propose solutions for extending $T_o$ to as long as 80. We focus on scaling relatively recent and continuous context lengths and not distant memory via language tokens \cite{sridhar2025memerscalingmemoryrobot, torne2026memmultiscaleembodiedmemory}. 

\subsection{Diffusion Policy Architecture}
\label{preliminaries:diffusion_architecture}
As depicted in Figure~\ref{fig:anchor_figure}, Diffusion Policy architecture comprises two stages: observation encoder to convert images to embeddings and backbone to denoise clean actions conditioned on embeddings and other tokens (proprioception, diffusion timestep, etc.). We present details for different architectures in Appendix~\ref{appx:diffusion_policy_details}. Note that majority of single-task models from Diffusion Policy \cite{diffusion_policy} and works building on it \cite{cadene2024lerobot, wei2025empiricalanalysissimandrealcotraining, robomimic2021, ren2024diffusionpolicypolicyoptimization} use a \textbf{UNet} \cite{ronneberger2015unetconvolutionalnetworksbiomedical} as the denoising backbone and merge conditioning via FiLM \cite{perez2017filmvisualreasoninggeneral}. \citet{torne2025learninglongcontextdiffusionpolicies} use \textbf{DiT} implementation of the original work \cite{diffusion_policy}.

\section{Evaluating Long-Context Diffusion Policies}
\label{sec:performance_trends}
We begin by investigating naive context length scaling under data and task variations. We train Diffusion Policies across \textbf{a)} context lengths ranging from short to long (specific values chosen per task), \textbf{b)} on tasks spanning memory requirements and with different local manipulation stability, and \textbf{c)} at three data scales per task. We present interpretable trends: long-context performance improves with data, and the scale needed for success significantly depends on simplicity of the manipulation primitive used in the tasks we evaluate. Moreover, naive scaling does not always fail. 

Throughout this section, we use \textbf{UNet + Cross Attention} architecture. The justification for choosing this as well as commentary on comparison between architectures is provided in Section~\ref{sec:cross_attention_conditioning}. We train diffusion policy for the following two categories of tasks (task details are in Appendix~\ref{appx:task_details}): 
\begin{enumerate}[leftmargin=1.5em, itemsep=1pt, topsep=0pt, parsep=0pt]
    \item \textbf{Tasks solvable with short context length: }We use the version of \textbf{push-T} from \cite{wei2025empiricalanalysissimandrealcotraining} and the \textbf{robomimic square} and \textbf{lift} tasks from \cite{robomimic2021}. These tasks show high success rate with short context lengths \cite{diffusion_policy, wei2025empiricalanalysissimandrealcotraining}. We study the performance at $T_o \in \{1, 2, 5, 10, 12, 16, 20\}$. 
    \item \textbf{Tasks requiring long context length: }We introduce two tasks, \textbf{push-and-return} and \textbf{grasp-and-return} (Section~\ref{subsec:task_and_success_criterion}). We train policies at $T_o \in \{4, 8, 32, 48, 60, 72, 80 \}$ for \textbf{push-and-return} and at $T_o \in \{4, 8, 16, 20, 24, 48 \}$ for \textbf{grasp-and-return}. We also provide limited experiments on \emph{hardware} at $T_o=92$ for \textbf{marshmallows} task adapted from \citet{mark2026bpplongcontextrobotimitation}. 
\end{enumerate}

\subsection{Benchmark Tasks Requiring Long Context Lengths}
\label{subsec:task_and_success_criterion}
\begin{wrapfigure}[16]{r}{0.45\textwidth}
    \centering
    \includegraphics[width=0.39\textwidth]{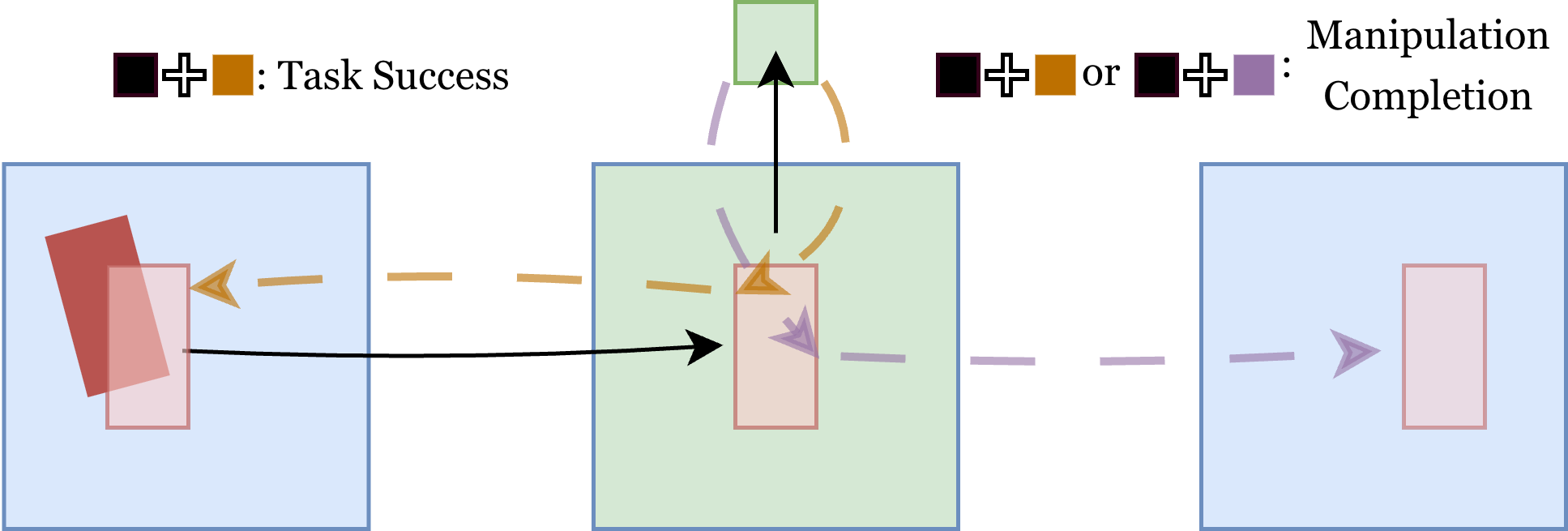}
    \caption{Schematic of the \textbf{push/grasp-and-return} tasks with two modes. A block spawns on either side. For \textbf{task success}, the robot should push/grasp the block to the center, move to a neutral location, and then return it to the area it originated in. Relaxing the memory requirement, \textbf{manipulation completion} is awarded for returning the block to either of the spawn locations. We define \textbf{contextual success} = $\frac{\mathbf{\text{task success}}}{\mathbf{\text{manipulation completion}}}$ to evaluate the percentage of rollouts showing manipulation completion that are also able to keep track of memory. } 
    \label{fig:push_or_grasp_and_return}
\end{wrapfigure}

The following tasks share the same goal structure but differ in the manipulation primitive used: 
\begin{enumerate}[leftmargin=1.5em, itemsep=1pt, topsep=0pt, parsep=0pt]
    \item \textbf{Push-and-Return: }A block spawns at one of the possible locations (Figure~\ref{fig:push_or_grasp_and_return}), and the robot must \textbf{1)} push it to the center, \textbf{2)} move to a neutral configuration, and \textbf{3)} push it back to the original spawn location. This task combines locally unstable planar pushing with memory dependent decisions about return location.  
    \item \textbf{Grasp-and-Return: }Same pattern as described for \textbf{push-and-return}, but the robot is required to achieve the objective via grasping a brick as opposed to pushing. This task combines the locally stable, prehensile manipulation skill of grasping a brick with contextual complexity.  
\end{enumerate}

Success is awarded if the policy achieves the stated objective (\textbf{task success}). In addition, we define the following two criteria (see Figure~\ref{fig:push_or_grasp_and_return}):
\begin{itemize}[leftmargin=1.5em, itemsep=1pt, topsep=0pt, parsep=0pt]
    \item \textbf{Manipulation completion: }The robot moves the block to the center and returns it to \emph{any} of the possible spawn locations. This criterion measures success in manipulation requirements of the task if memory requirement is relaxed. 
    \item \textbf{Contextual success: }The percentage of \textbf{manipulation completion} rollouts which return the block/brick to the correct return location. 
\end{itemize}
\textbf{Marshmallows} task involves scooping marshmallows from a large bowl to a smaller target bowl exactly twice and then hitting a button to indicate task completion. Due to arbitrary number of initial marshmallows in the target bowl, long memory is the only way to keep track of success. 

\subsection{Empirical Results}
\label{subsec:empirical_validation_performance_trends}
For each task in simulation, we train at 3 data scales: $N/2$, $N$, and $2N$. The choice of $N$ is specific to the task: it is the amount of data that would typically be available for training single task policy. See Appendix~\ref{appx:task_details}. Overall, we find that naively scaling context length is not as brittle as might have been indicated in prior work \cite{torne2025learninglongcontextdiffusionpolicies, sridhar2025memerscalingmemoryrobot}. Performance at longer context lengths is task dependent, with interpretable trends. 

\begin{figure}[!tbp]
    \centering
    \begin{subfigure}[t]{0.31\textwidth}
        \centering
        \includegraphics[width=\textwidth]{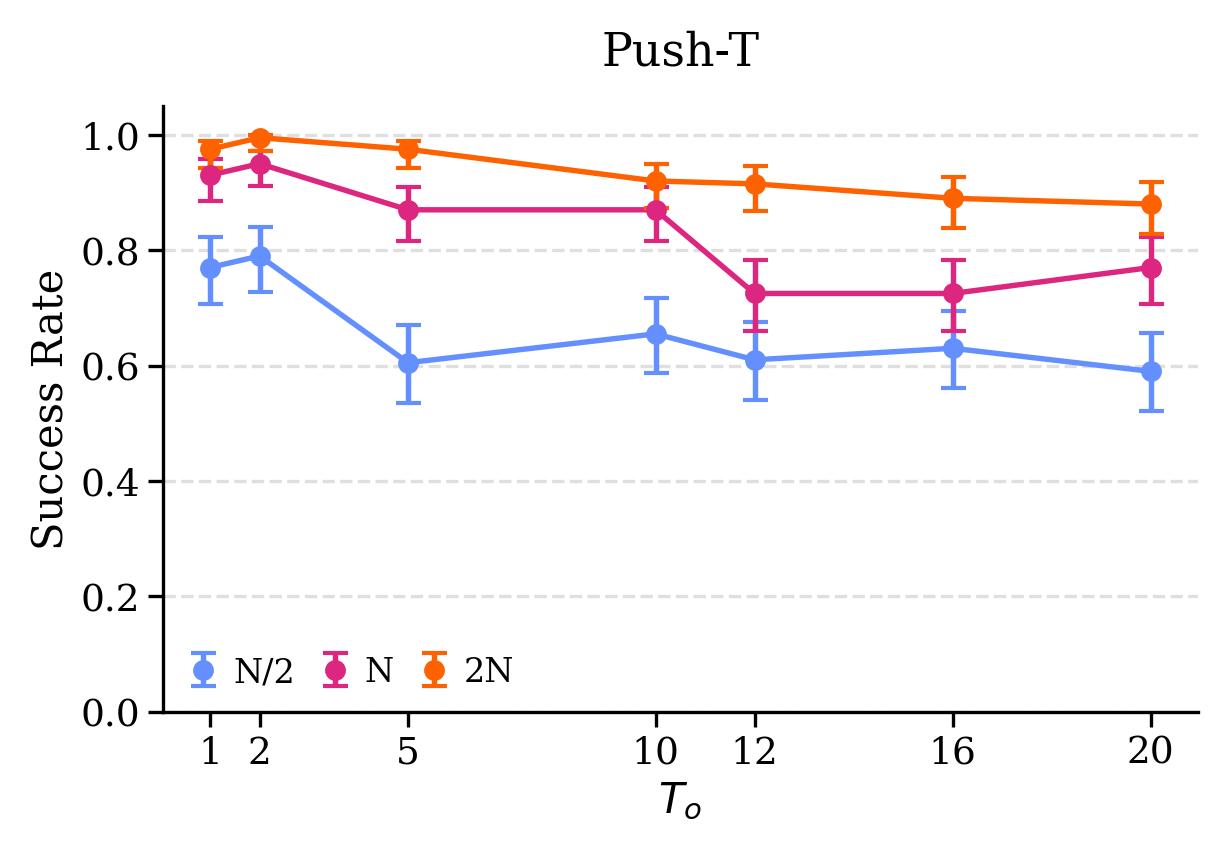}
        \caption{}
        \label{fig:push_t_data_cl_scaling}
    \end{subfigure}
    \hfill
    \begin{subfigure}[t]{0.31\textwidth}
        \centering
        \includegraphics[width=\textwidth]{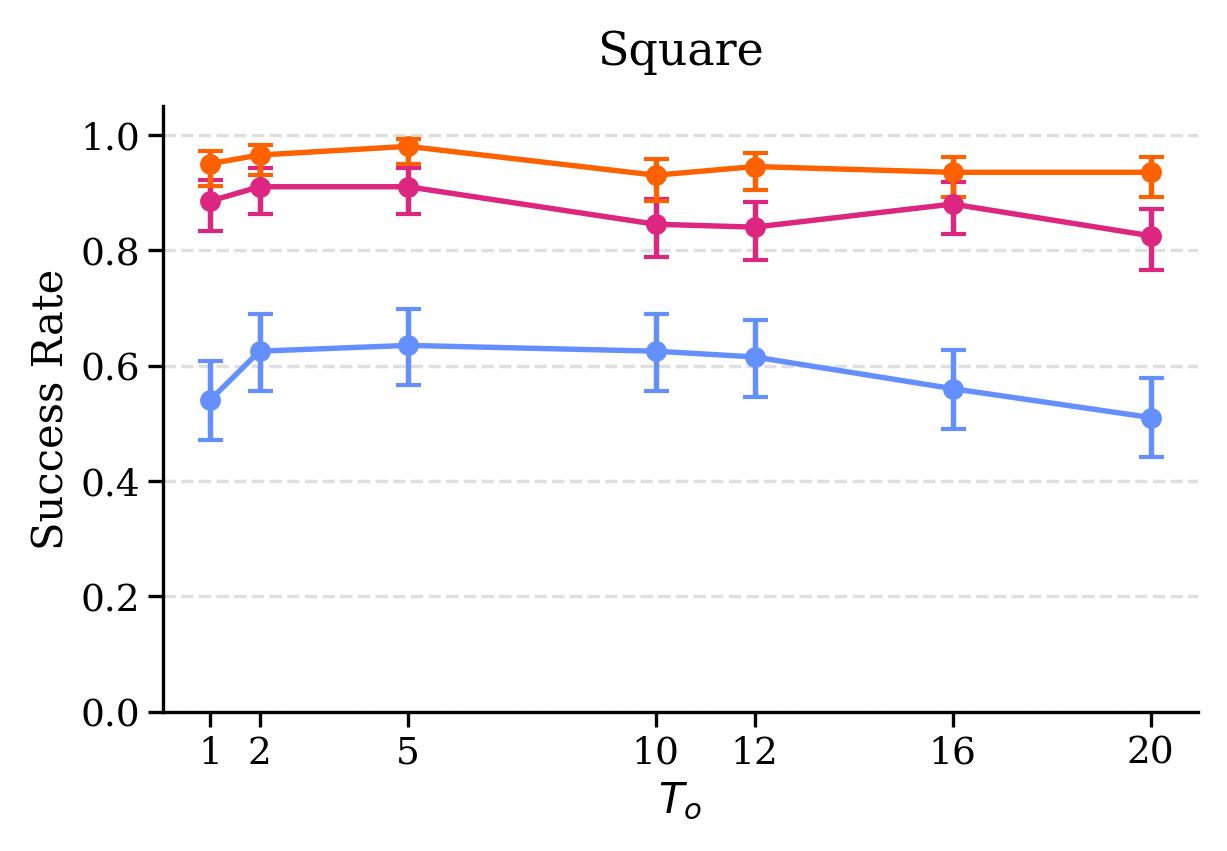}
        \caption{}
        \label{fig:square_data_cl_scaling}
    \end{subfigure}
    \hfill
    \begin{subfigure}[t]{0.31\textwidth}
        \centering
        \includegraphics[width=\textwidth]{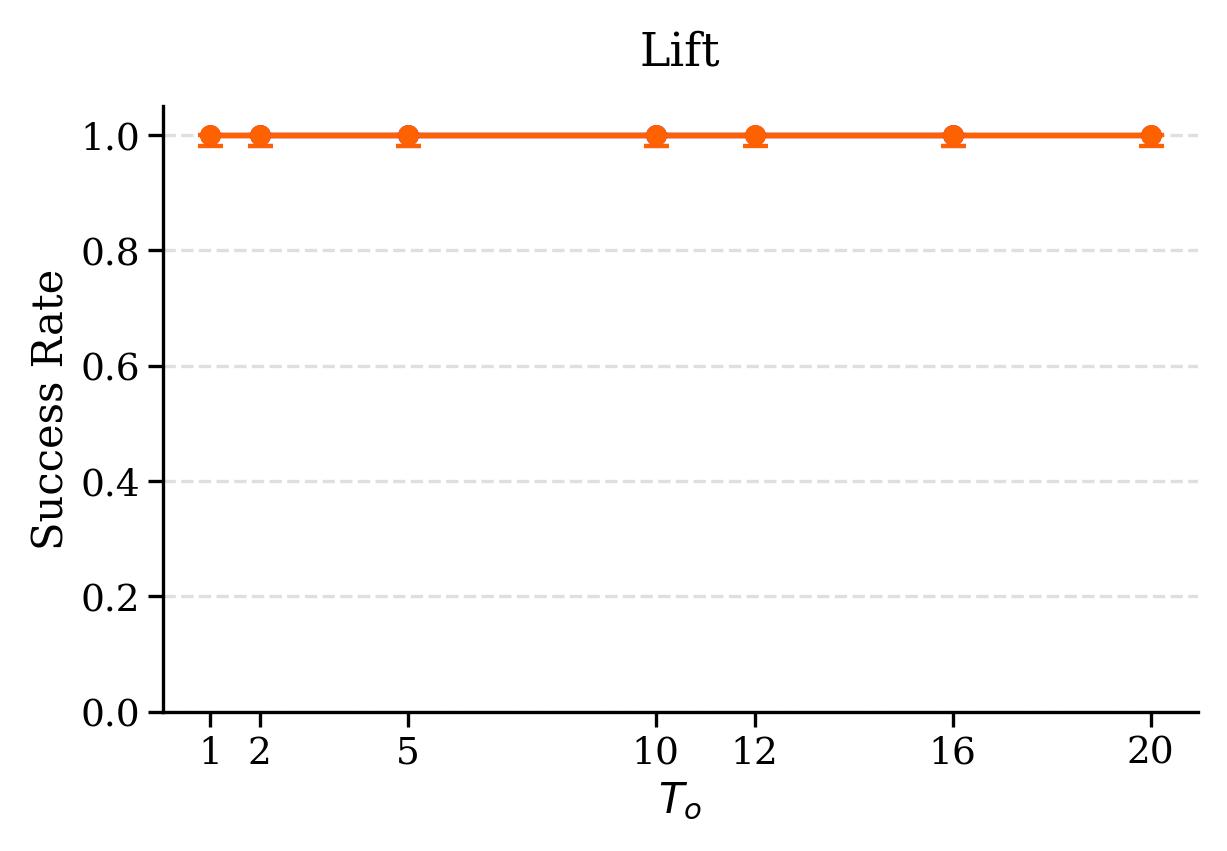}
        \caption{}
        \label{fig:lift_data_cl_scaling}
    \end{subfigure}
    \caption{\textbf{Short-context solvable tasks' performance.} Long-context policy performance improves as data is scaled. The difference between short and long-context policies shrinks in high data regime.}
    \label{fig:context_length_scaling_tasks}
\end{figure}

\textbf{Tasks solvable with short-context lengths: }Figure~\ref{fig:context_length_scaling_tasks} shows policy performance as context length is increased at different data scales for \textbf{push-T}, \textbf{square}, and \textbf{lift}. Regardless of data scale, \textbf{lift} shows no drop in performance. On the contrary, \textbf{square} and \textbf{push-T} performance decreases beyond short context lengths in the low data regime ($N/2$). The magnitude of drop, however, reduces as more data is added, with \textbf{square} showing statistically insignificant drop in the high data regime. These results give important insights: \textbf{a)} performance drop from short to long context lengths is highly task and data dependent \textbf{b)} with this architecture, the drop in performance for these tasks is minimal in the high data regime (in Section \ref{sec:cross_attention_conditioning} we show that the performance can drop for other architectures).

\begin{figure}[!tbp]
    \centering
    \begin{subfigure}[t]{0.32\textwidth}
        \centering
        \includegraphics[width=\textwidth]{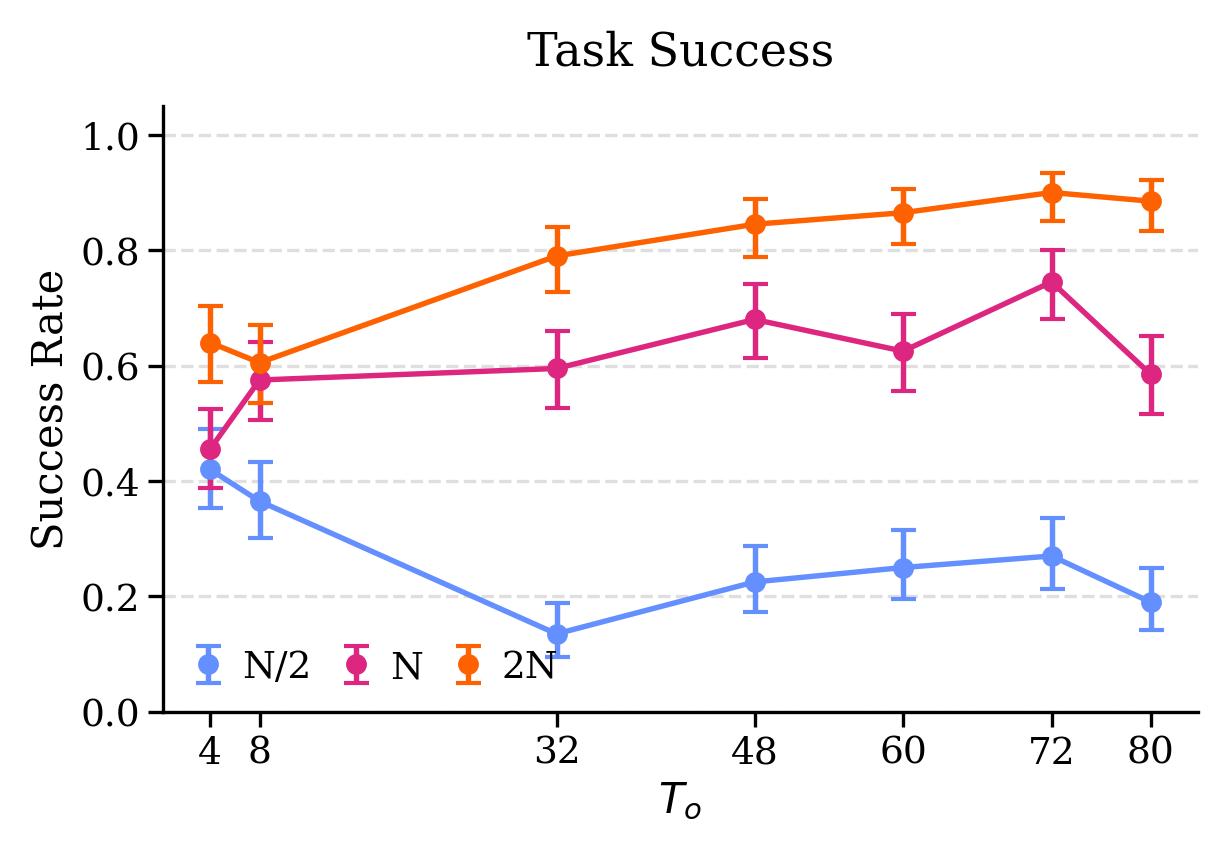}
        \caption{Even though the task requires memory, long-context policies perform worse than short context policies in the limited ($N/2$) data regime. In the $N$ and $2N$ data regimes, however, we notice the expected rise in performance.}
        \label{fig:push_and_return_cross_task_success_data_scaling}
    \end{subfigure}
    \hfill
    \begin{subfigure}[t]{0.32\textwidth}
        \centering
        \includegraphics[width=\textwidth]{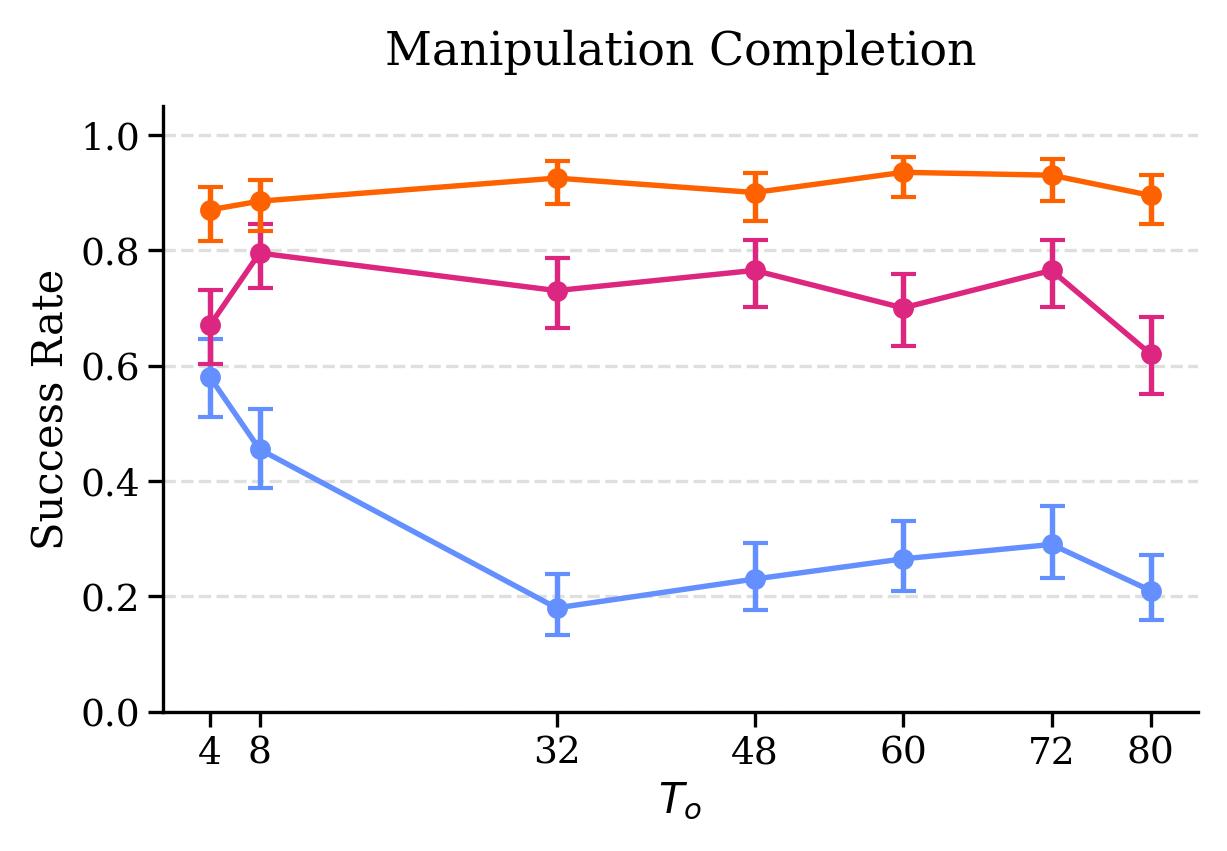}
        \caption{Long-context policies in the limited data regime aren't able to complete the task even if the memory requirement for success is relaxed. This changes as data is scaled. Short-context policies are successful regardless of data.}
        \label{fig:push_and_return_cross_subtask_success_data_scaling}
    \end{subfigure}
    \hfill
    \begin{subfigure}[t]{0.32\textwidth}
        \centering
        \includegraphics[width=\textwidth]{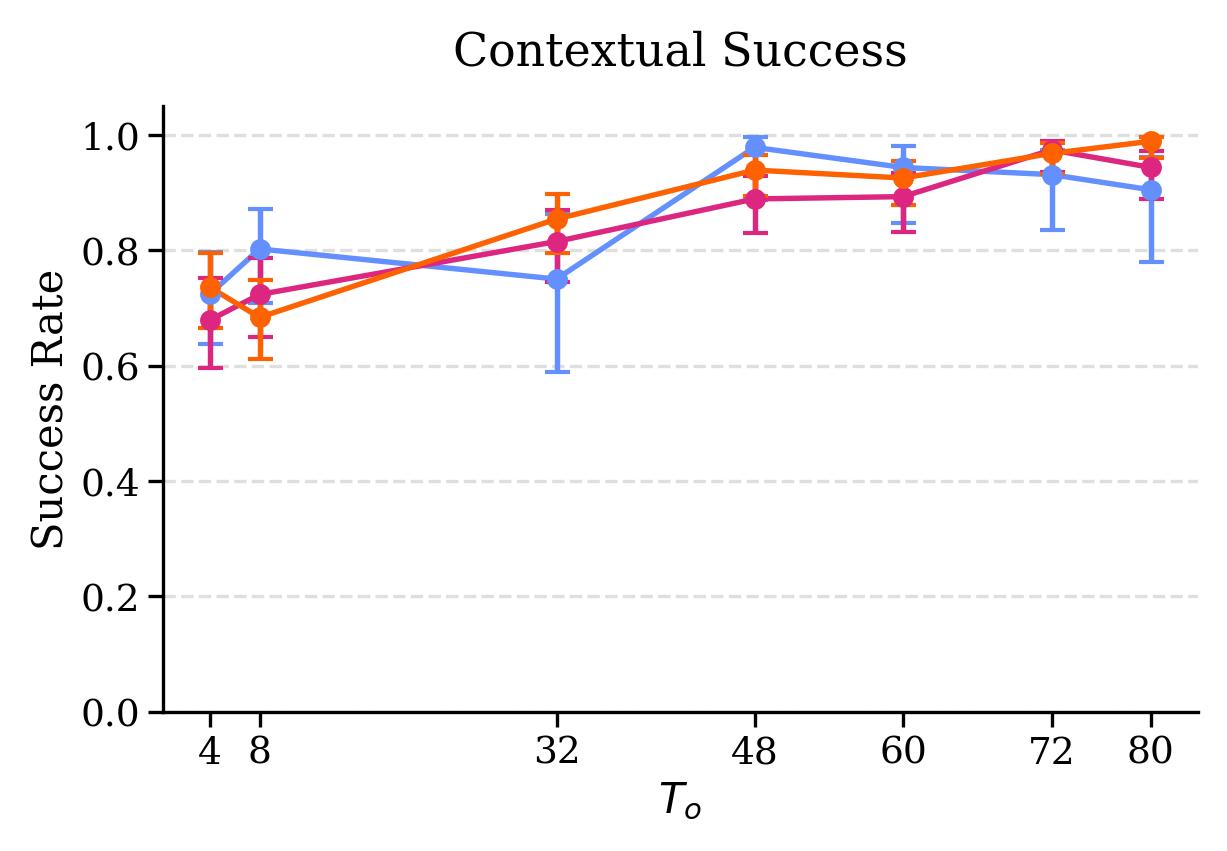}
        \caption{Long-context policies always show high contextual success. Thus, those long-context policies which are able to complete manipulation requirements also often keep track of history.}
        \label{fig:push_and_return_cross_contextual_success_data_scaling}
    \end{subfigure}
    \caption{\textbf{Push-and-return task performance. }}
    \label{fig:push_return_data_scaling}
\end{figure}

\begin{figure}[!tbp]
    \centering
    \begin{subfigure}[t]{0.31\textwidth}
        \centering
        \includegraphics[width=\textwidth]{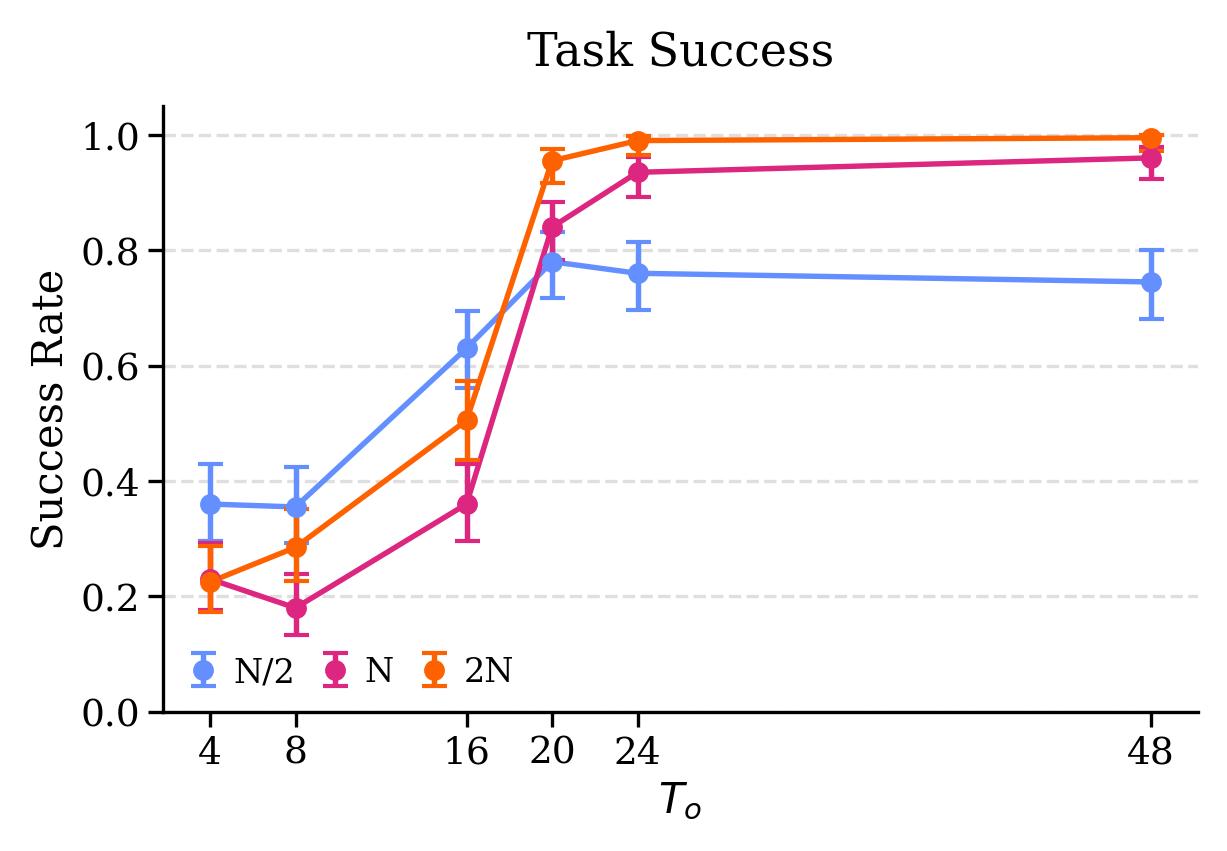}
        \caption{Long-context policies show reasonable performance even in the low data regime ($N/2$), which further improves as data is increased.}
        \label{fig:grasp_and_return_cross_task_success_data_scaling}
    \end{subfigure}
    \hfill
    \begin{subfigure}[t]{0.31\textwidth}
        \centering
        \includegraphics[width=\textwidth]{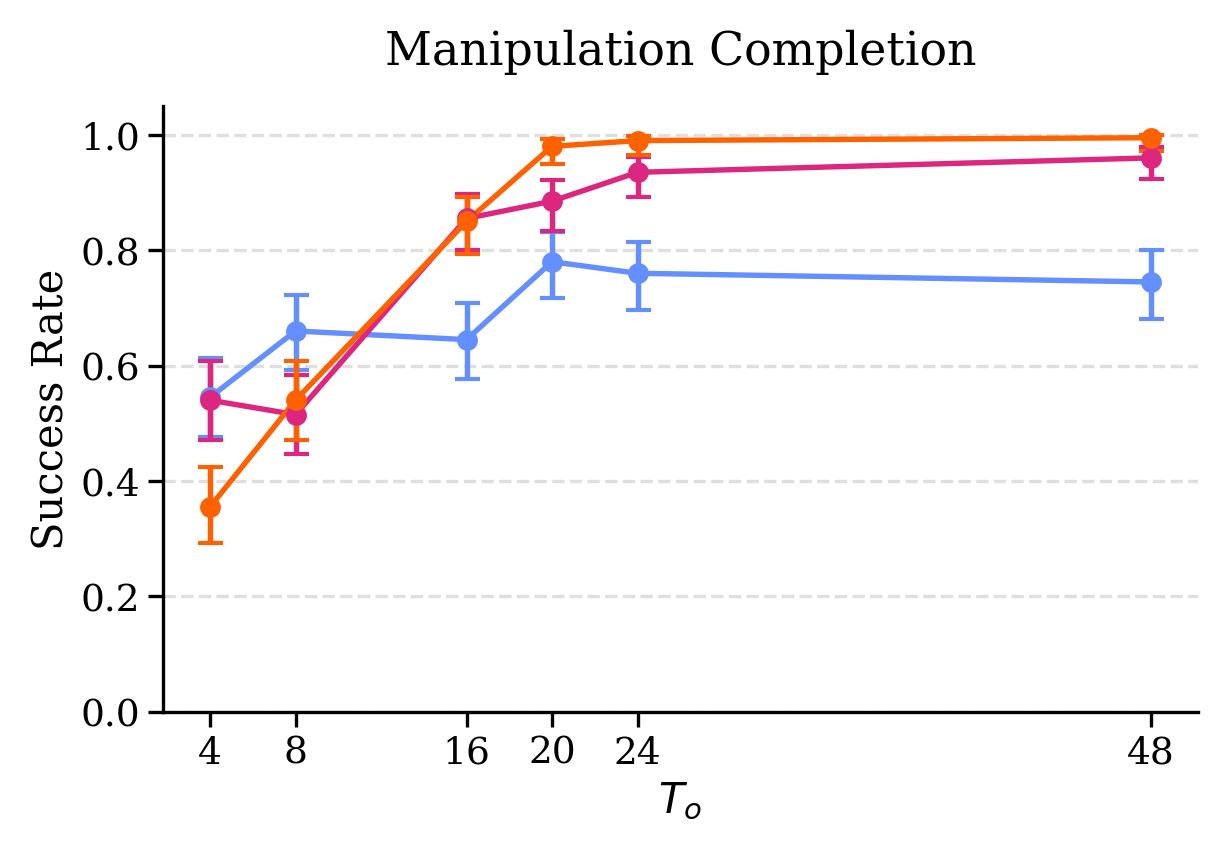}
        \caption{Long-context policies are always able to reasonably complete manipulation requirements. Contrary to \textbf{push-and-return}, however, short-context policies are unable to do so regardless of data.}
        \label{fig:grasp_and_return_cross_subtask_success_data_scaling}
    \end{subfigure}
    \hfill
    \begin{subfigure}[t]{0.31\textwidth}
        \centering
        \includegraphics[width=\textwidth]{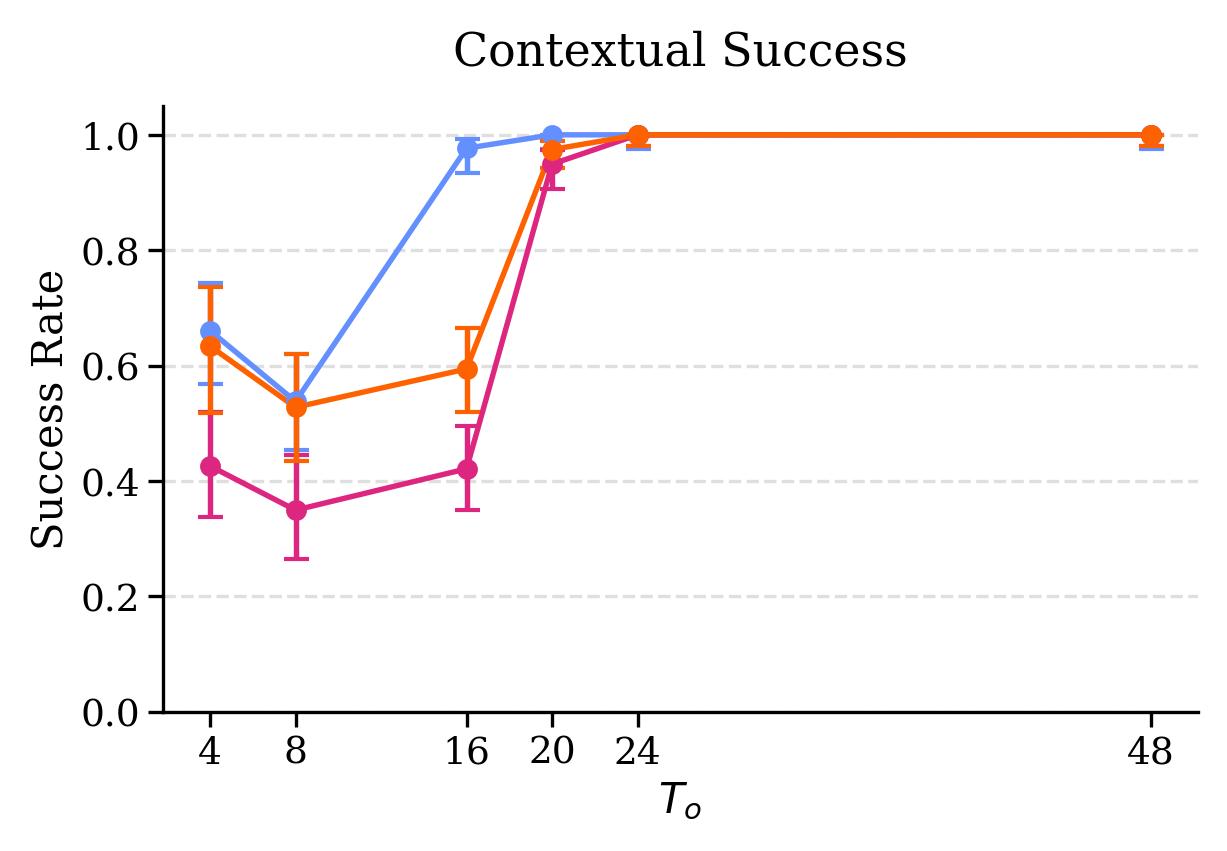}
        \caption{Long-context policies always show high contextual success. Thus, those long-context policies which are able to complete manipulation requirements also often keep track of history.}
        \label{fig:grasp_and_return_cross_contextual_success_data_scaling}
    \end{subfigure}
    \caption{\textbf{Grasp-and-return task performance.}}
    \label{fig:grasp_return_data_scaling}
\end{figure}

\pagebreak 

\begin{wrapfigure}[15]{r}{0.45\textwidth}
    \centering
    \includegraphics[width=0.39\textwidth]{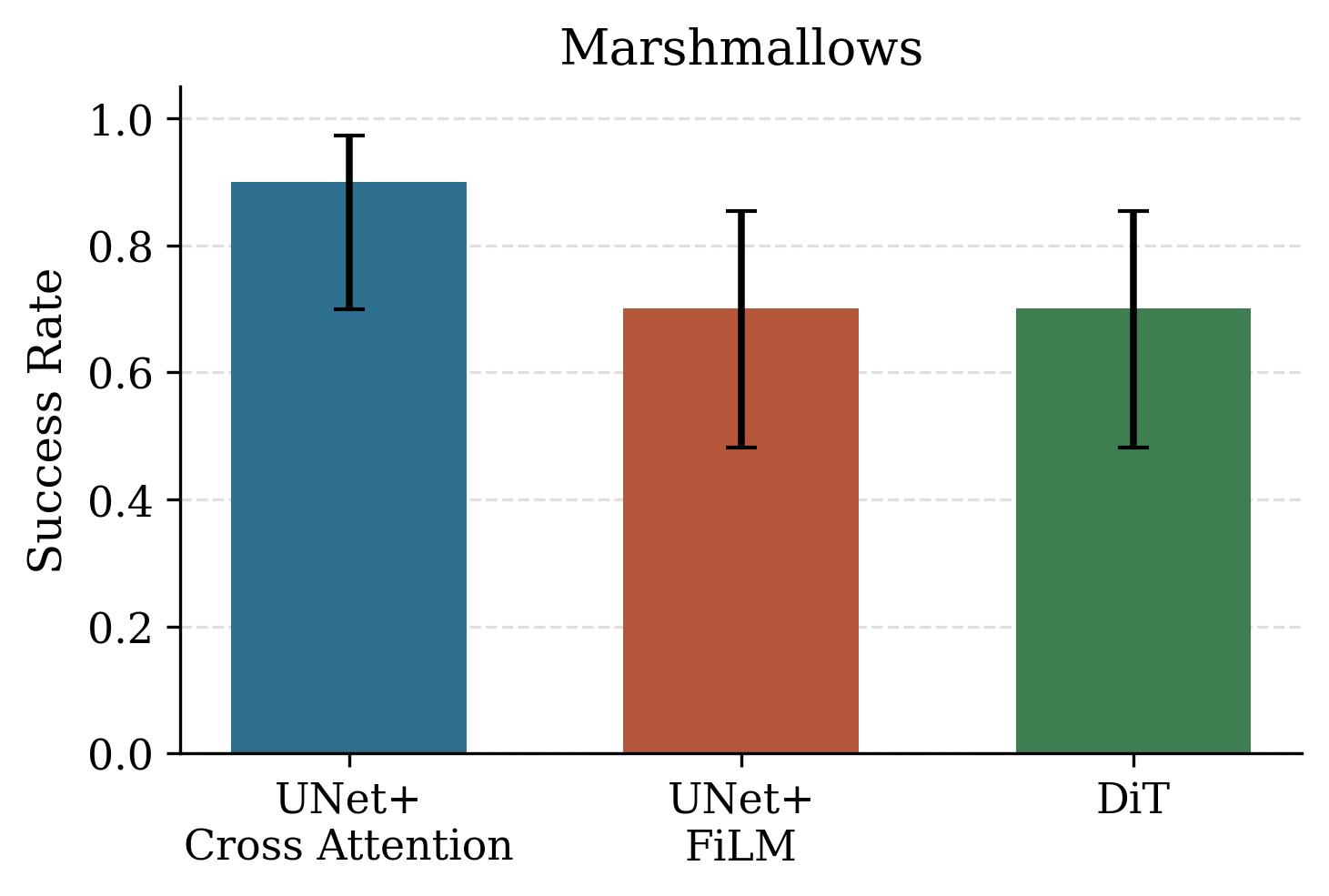}
    \caption{Results for the \textbf{marshmallows} task, where through simple hyperparameter changes we are able to achieve high performance with just 100 training trajectories and a naively scaled context length of $T_o=92$.} 
    \label{fig:hardware_marshmallows}
\end{wrapfigure}

\textbf{Tasks requiring long context lengths: } As shown in Figures~\ref{fig:push_and_return_cross_task_success_data_scaling} and \ref{fig:grasp_and_return_cross_task_success_data_scaling}, \textbf{task success} significantly improves for both \textbf{push-and-return} and \textbf{grasp-and-return} tasks as context length is increased. The primary difference is about data: \textbf{grasp-and-return} shows good long-context performance with just $N/2$ trajectories, whereas acceptable performance for \textbf{push-and-return} is $N$ trajectories onward. 

Additionally, we show that it is, in fact, possible to get successful long-context policies on the \textbf{marshmallows} task through naive scaling (contrary to what is noted in \citet{mark2026bpplongcontextrobotimitation}, though difference in embodiment and environment could explain this). Figure~\ref{fig:hardware_marshmallows} shows that with simple architectural and hyperparameter changes, we are able to achieve high success rate on the task. We defer detailed commentary to Appendix~\ref{appx:hardware_experiments}. 

Collectively, these results show that success at longer context lengths is strongly data scale dependent. Sample complexity for good performance with long context lengths seems to be dictated, to a significant extent, by complexity of the manipulation primitive used: locally stable prehensile tasks like \textbf{lift} and \textbf{grasp-and-return} show favorable trends even in the low data regime, whereas tasks like \textbf{push-T}, \textbf{square}, and \textbf{push-and-return} require more data. This is consistent with the intuition frequently used for short-context policy learning. We demonstrate the favorable impact of locally stable prehensile manipulation in Appendix~\ref{appx:grasp_as_manipulation_primitive}.

It is reasonable to ask: \emph{why do prior works strongly criticize naive context length scaling?} We attribute this to absence of benchmarking on dataset size, architecture choice, and task properties. We provide detailed discussion in Appendix~\ref{appx:naive_scaling_criticism}. 

\subsection{Additional Comments}
Investigating Figures~\ref{fig:push_and_return_cross_subtask_success_data_scaling} and \ref{fig:grasp_and_return_cross_subtask_success_data_scaling} for \textbf{push-and-return} and \textbf{grasp-and-return} respectively, we note that \textbf{manipulation completion} at longer context lengths for both tasks mimics the trends in \textbf{task success}. Further investigating Figures~\ref{fig:push_and_return_cross_contextual_success_data_scaling} and \ref{fig:grasp_and_return_cross_contextual_success_data_scaling} for \textbf{contextual success} in these tasks respectively, we find that those long-context policies which complete manipulation often also manage to keep track of contextual requirements. Thus, we hypothesize that when long-context policies do fail, it is primarily due to failure at learning to manipulate within the domain, not due to failure to keep track of requirements arising from history. Further discussion is in Appendix~\ref{appx:reasons_for_long_context_failure}.  

Another surprising part of the \textbf{manipulation completion} criterion is seen at short context lengths ($T_o \in \{4, 8 \}$). While short-context \textbf{manipulation completion}, as expected, is high for \textbf{push-and-return} at all data scales, it is low for \textbf{grasp-and-return} regardless of data scale. We attribute this to observability of expert hidden state and defer discussion to Appendix~\ref{appx:when_is_short_context_sufficient}. 

\section{Training Long-Context Diffusion Policies}
\label{sec:lc_policy_learning}
While scaling data helps with long-context learning (Section~\ref{sec:performance_trends}), we now present solutions for the \emph{limited data regime}. We begin by establishing \textbf{UNet} as a favorable action denoising backbone choice when training single-task diffusion policies from scratch, with \textbf{cross-attention} a favorable conditioning method in low data regime (Section~\ref{sec:cross_attention_conditioning}). We then propose a new learning algorithm for long-context policies (Section~\ref{sec:variable_history_training}), and revisit prior work in long-context learning \cite{torne2025learninglongcontextdiffusionpolicies} (Section~\ref{subsec:revisiting_ptp}).

\subsection{Cross-Attention Conditioning}
\label{sec:cross_attention_conditioning}
We compare the frequently used \textbf{UNet + FiLM} and \textbf{DiT} architectures with \textbf{UNet + Cross-Attention} on \textbf{push-T}, \textbf{square}, and \textbf{push-and-return}. While cross-attention conditioning for UNets has been widely used in the computer vision community before \cite{rombach2022highresolutionimagesynthesislatent}, we have not seen it used to train robot learning policies. A detailed design explanation of the 3 diffusion policy architectures compared in this section is presented in Appendix~\ref{appx:dp_architecture_comparison}. Figure~\ref{fig:unet_cross_vs_film_all_tasks_combined} shows the comparison in the $N/2$ and $N$ data regimes.

\begin{figure}[!tbp]
    \centering
\includegraphics[width=0.9\linewidth,height=7.8cm,keepaspectratio]{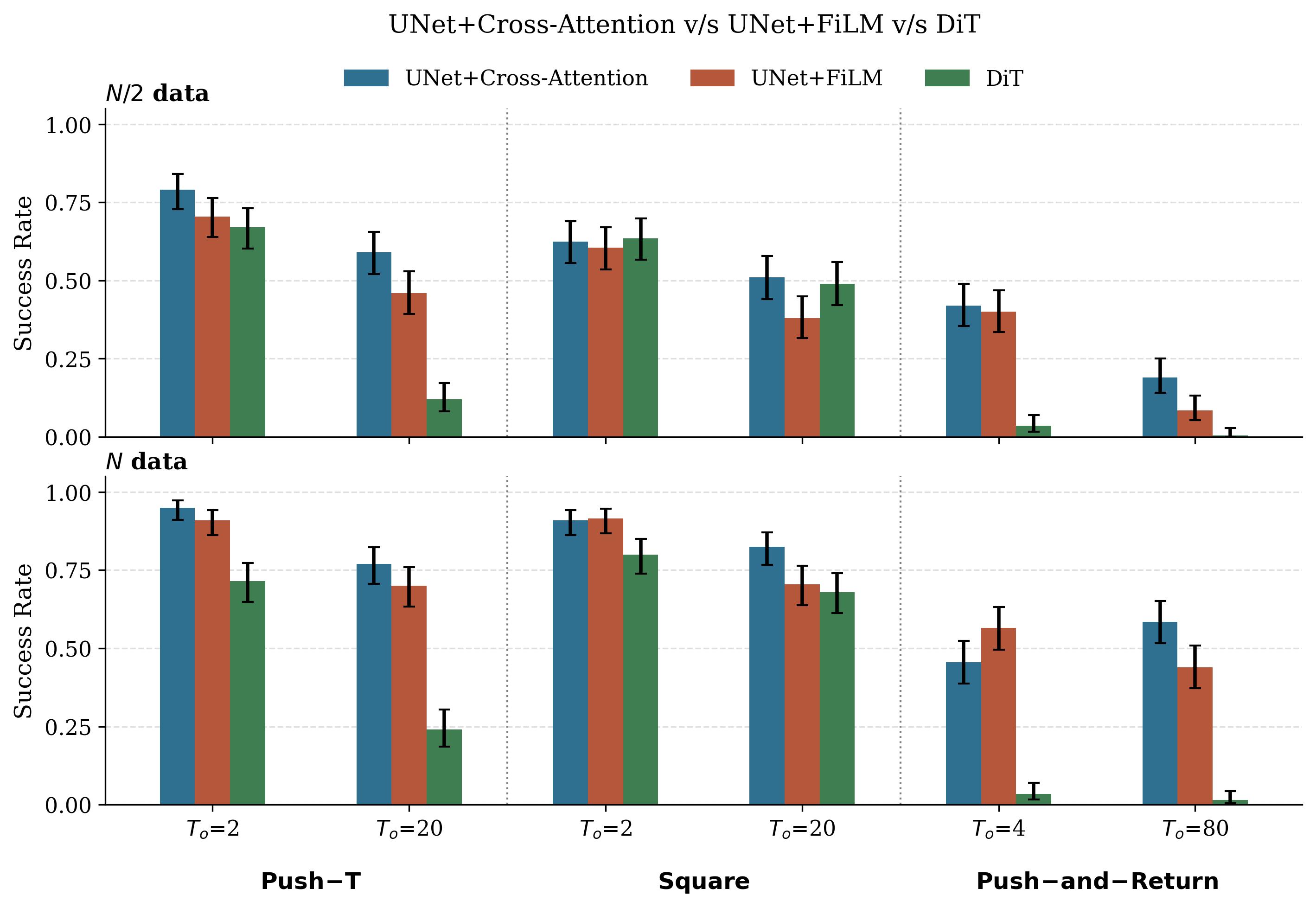}
    \caption{Comparison of \textbf{UNet + Cross Attention}, \textbf{UNet + FiLM} and \textbf{DiT}. \textbf{DiT} shows catastrophic failure in significant number of cases. The other two architectures perform similarly at short context lengths, while \textbf{UNet + Cross Attention} shows improvement at longer context lengths especially in the limited data regime. Figure~\ref{fig:unet_cross_vs_film_all_tasks} shows comparison by data scale at a given context length. As data is scaled, performance gap between architectures decreases, suggesting that cross-attention provides an inductive bias that reduces the sample complexity of long-context policy learning.\vspace{-\baselineskip}}
    \label{fig:unet_cross_vs_film_all_tasks_combined}
\end{figure}

The \textbf{DiT} architecture from \citet{diffusion_policy} largely fails, even showing catastrophic failure in \textbf{push-and-return}. In fact, the behavior of \textbf{DiT} noticed for \textbf{push-T} when comparing performance at $T_o=2$ and $T_o=20$ is similar to the trends noted by \citet{torne2025learninglongcontextdiffusionpolicies} and could explain their criticism of naive scaling. This architecture should clearly not be chosen as a baseline.

The \textbf{UNet+Cross-Attention} architecture provides an advantage in the low data regime at long context lengths. However, at short context lengths, the performance, especially between the two UNet architectures, is relatively similar. This could be the reason why most robot learning works using short-context diffusion policy have not looked beyond \textbf{UNet+FiLM} \cite{cadene2024lerobot, wei2025empiricalanalysissimandrealcotraining, robomimic2021, ren2024diffusionpolicypolicyoptimization}. Further discussion and comparison organized by data regime at a given context length is presented in Appendix~\ref{appx:cross_atten_conditioning_additional}. 

\subsection{Variable History Training}
\label{sec:variable_history_training}
While architectural inductive bias (Section~\ref{sec:cross_attention_conditioning}) shows improvements, we want to further close the performance gap between short and long context policies as noted in the $N/2$ data regime. Given our investigation in Appendix~\ref{appx:reasons_for_long_context_failure} and supported by prior literature \cite{mark2026bpplongcontextrobotimitation, dehaan2019causalconfusionimitationlearning}, we speculate that long-context policies perform worse because of over-fitting with limited data, which then leads to worse co-variate shift when deployed closed loop. Evidently, short-context policies do not exhibit this. We therefore propose a joint history training method, where a model learns over a curriculum of context lengths as opposed to just one fixed context length, using the short-context datagrams to mitigate over-fitting while still benefiting from memory when useful (and ignoring when not).

For a discrete sample space $\mathcal{C}$, let $\Delta (\mathcal{C})$ denote the set of all possible probability distributions over $\mathcal{C}$. Our learning algorithm is described in Algorithm~\ref{alg:diffusion_vh_training}.  

\begin{algorithm}[h]
\caption{Variable History Training for Diffusion Policy}
\label{alg:diffusion_vh_training}
\begin{algorithmic}[1]
\Require Dataset $\mathcal{D} = \{\tau_1, \tau_2, \dots \tau_N \}, \quad \tau_i = \{(o_1, a_1), (o_2, a_2), \dots (o_{T_i}, a_{T_i}) \}$, a minimum context length $c \in \{1, \dots, T_o\}$, a training step varying distribution over $\mathcal{C} = \{c, c+1, \dots, T_o\}$ denoted by $\rho_i \in \Delta(\mathcal{C})$, where $i$ is the training step. Define $T_p^{past}(m) = \min \{m, T_p^{past}\}$
\Ensure Trained model parameters $\theta$
\State Initialize $\theta$ randomly
\For{each training iteration $i$}
    \State Sample datagrams from $\mathcal{D}$ and $m \sim \rho_i$
    \State $\mathcal{L} \gets \| \pi_\theta(\mathbf{o}[k-m+1:k]) - \mathbf{a}[k - T^{past}_p(m) + 1:k+T_p] \|^2$
    \State Update $\theta$ via gradient descent on $\mathcal{L}$
\EndFor
\State \Return $\theta$
\end{algorithmic}
\end{algorithm}

The key hyperparameters in Algorithm~\ref{alg:diffusion_vh_training} are past prediction horizon $T^{past}_p$ and training step varying distribution $\rho_i$. We try the following for $\rho_i$: 
\begin{enumerate}[leftmargin=1.5em, itemsep=1pt, topsep=0pt, parsep=0pt]
    \item \textbf{Random Sprinkle}: Assign a relatively large probability $p$ to $T_o$ and sample from $\mathcal{U}(c, c+1, \dots, T_o-1)$ with probability $1-p$. In our experiments, we set $p=0.8$. In this case, $\rho_i$ is constant across $i$. 
    \item \textbf{Progressive}: Split $\mathcal{C}$ into $\mathcal{C}_1$ and $\mathcal{C}_2 = \mathcal{C} \setminus \mathcal{C}_1$. Start with $\mathcal{C}_1 = \{c\}$, and $\rho_0 = \mathcal{U}(\mathcal{C}_1)$. Up to half the total training steps, sequentially move an element from $\mathcal{C}_2$ to $\mathcal{C}_1$ and $\rho_i = \mathcal{U}(\mathcal{C}_1)$. Once $\mathcal{C}_2$ is empty, change to \textbf{random sprinkle} method for the remaining steps. 
\end{enumerate}

While training on single context length, we use \textbf{full} past prediction: $T^{past}_p = T_o$ (as recommended by the code from Diffusion Policy \cite{diffusion_policy}). In the variable history training with limited data case, however, we notice \textbf{short} past prediction ($T^{past}_p$ is set to a small value compared to $T_o$) works better. Our ablations from Appendix~\ref{appx:vht_ablations} show us that when data regime is insufficient for long-context learning, \textbf{progressive} is a better choice. However, as data is scaled and naive scaling is good enough, \textbf{random sprinkle} with \textbf{full} past prediction is a better curriculum choice which maintains performance compared to naive scaling. 

\begin{figure}[!tbp]
    \centering
    \includegraphics[width=0.95\linewidth]{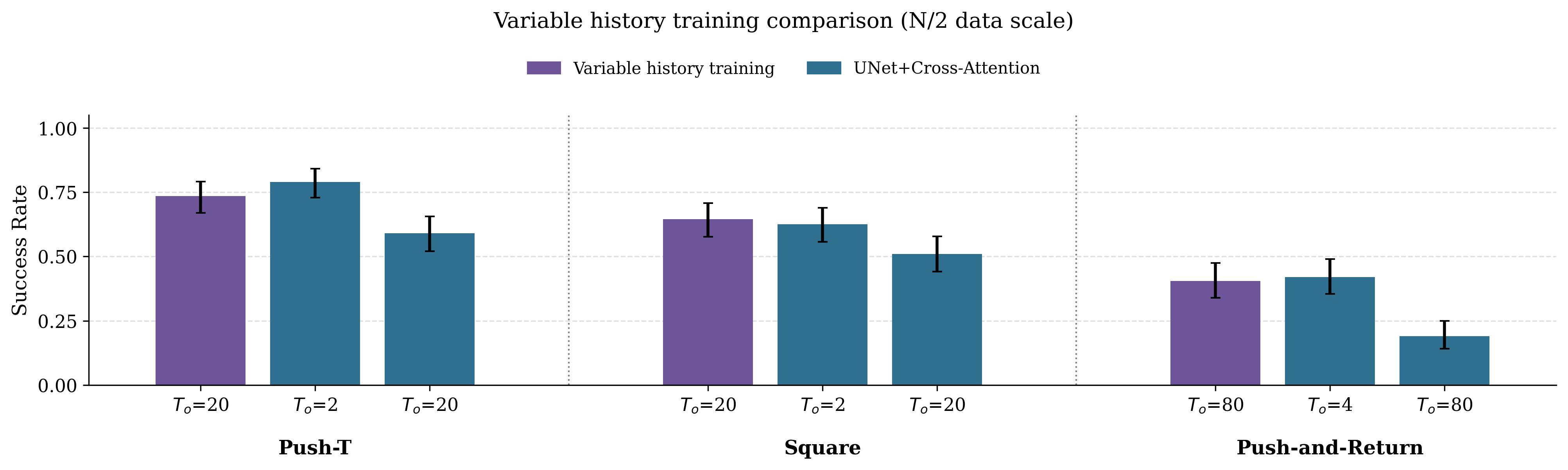}
    \caption{In the low ($N/2$) data regime, variable history training leads to gains in long-context performance. Policies are able to perform as well as the best short-context policy. See Figure~\ref{fig:variable_vs_cross} for $N$ data regime performance. \vspace{-\baselineskip}}
    \label{fig:variable_vs_cross_combined}
\end{figure}

We present results for our algorithm in the low ($N/2$) data regime in Figure~\ref{fig:variable_vs_cross_combined}. We notice that with our variable training method, long-context policies, even in the low data regime, match the performance of the short-context policy (and show significant improvements over simply naively scaling to long-context). Figure~\ref{fig:variable_vs_cross} shows that when data is sufficient for good long-context policy performance with naive scaling ($N$ data regime), the method does not lead to drop in performance. It can thus be used in a general manner on tasks without requiring benchmarking for dataset size. 

\subsection{Revisiting Past-Token Prediction}
\label{subsec:revisiting_ptp}
\begin{figure}[!tbp]
    \centering
    \includegraphics[width=0.95\linewidth]{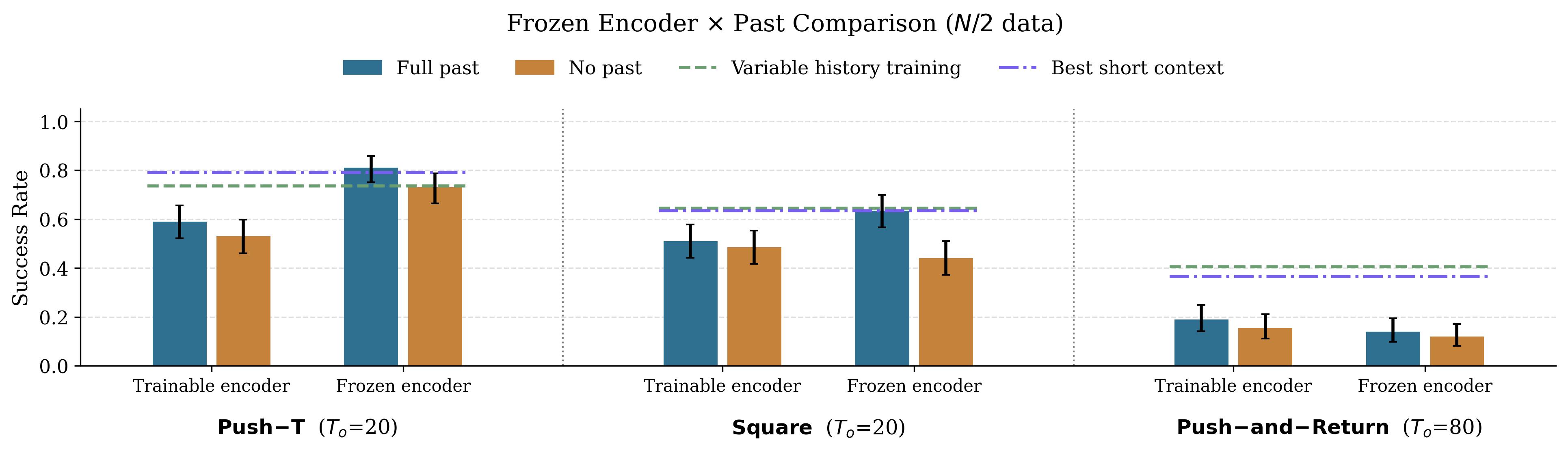}
    \caption{Results comparing trainable encoder and encoder frozen from short context lengths with past action prediction. Past-action prediction and frozen observation encoder seem to be driving each others' success, but \textbf{push-and-return} is an exception. \vspace{-\baselineskip}}
    \label{fig:frozen_encoder_past_comparison_N_half}
\end{figure}

\citet{torne2025learninglongcontextdiffusionpolicies} proposed predicting past actions as an auxiliary loss that helps with long-context learning. All our naive scaling policies from Sections~\ref{sec:performance_trends} and \ref{sec:cross_attention_conditioning} use past action prediction (a design choice from \cite{diffusion_policy}). Yet, we do not always see successful/improving long-context policy performance. In addition to past-token prediction, they make other design choices, such as freezing observation encoder from short-context policy, purely as a practical training optimization technique claiming no impact on policy performance. In Figure~\ref{fig:frozen_encoder_past_comparison_N_half}, we explicitly compare which aspect helps, and find that encoder freezing is certainly contributing to success, either independently or in conjunction with past-action prediction. However, there are exceptions (\textbf{push-and-return}), and our method matches/out-performs. Note that our implementation of past-token prediction differs slightly from that of \citet{torne2025learninglongcontextdiffusionpolicies}; we use our suggested \textbf{UNet+Cross-Attention} baseline, as the \textbf{DiT} architecture they use shows poor performance even with favorable hyperparameter changes (Figure~\ref{fig:unet_cross_vs_film_all_tasks_combined}). We defer further discussion to Appendix~\ref{appx:past-token_prediction}. 
 
\section{Limitations and Discussion}
\label{sec:limitations}
While our work is about model architecture, inference time overhead due to longer context length remains a challenge. For instance, distillation methods to lightweight architectures could be improved to account for long-context policies, speeding up inference despite a large set of image inputs that need to be processed by the policy in closed loop. 

Further, while we present variable histories at training time, we hope the algorithm encourages the robotics community to think of robot policies closer to the realm of agentic LLMs. A policy trained on multiple context lengths can also use variable context lengths at inference time, and algorithms establishing how to regulate inference time context length input are a promising future direction. 

\section{Conclusion}
\label{sec:conclusion}
Varying data scale and task properties, we show that naively scaling context length is not necessarily catastrophic. We have established sensitivity to architecture in long-context imitation learning, showing  UNet+Cross-Attention provides a helpful inductive bias when training single-task policies from scratch in the low data regime. We show that through variable history training, long-context policy performance is improved in the low data regime and maintained otherwise. Finally, we have commented on the underlying mechanisms of past-action prediction as an auxiliary loss \cite{torne2025learninglongcontextdiffusionpolicies}.

\acknowledgments{This material is based upon work supported by the National Science Foundation (NSF) Engineering Research Center for Human AugmentatioN via Dexterity (HAND) (Grant No. 2330040), NSF Graduate Research Fellowship Program under Grant No. DGE-2141064, Natural Sciences and Engineering Research Council of
Canada CGS D-587703, and Amazon.com Services LLC PO \#2D-19307345. Any opinions, findings, and conclusions or recommendations expressed in this material are those of the author(s) only. 

We thank The Infrastructure Group at MIT CSAIL for maintaining the compute resources primarily used for this work. Additionally, this work used ACES computing resources provided by Texas A\&M High Performance Research Computing through allocation CIS250858 from the Advanced Cyberinfrastructure Coordination Ecosystem: Services \& Support (ACCESS) program, which is supported by U.S. NSF grants \#2138259, \#2138286, \#2138307, \#2137603, and \#2138296 \cite{nsfaccess}. We also acknowledge the MIT SuperCloud and Lincoln Laboratory Supercomputing Center for providing HPC, database, and consultation resources \cite{reuther2018interactive}.}

\bibliography{ref}  %

\appendix

\section{Additional Related Works}
\label{appx:additional_related_works}
\subsection{Context Length in Imitation Learning}
We have already introduced and discussed the work from \citet{torne2025learninglongcontextdiffusionpolicies, mark2026bpplongcontextrobotimitation, sridhar2025memerscalingmemoryrobot}. We now elaborate on these. \citet{mark2026bpplongcontextrobotimitation} suggest using a pre-trained VLM to identify key events in policy history, and only conditioning the policy on these VLM identified frames. They also argue that naively increasing context length can avoid failure, but focus on action chunking as the reason for this. They have limited data scaling ablation, but it largely agrees with our results. Other methods use language based summarization of observation histories \cite{sridhar2025memerscalingmemoryrobot, torne2026memmultiscaleembodiedmemory}. They keep track of historic events through language/intermediate tokens summarizing history of observations. While \cite{sridhar2025memerscalingmemoryrobot} rely on just the most recent observation for control, \cite{torne2026memmultiscaleembodiedmemory} introduce video based observation encoding for the recent history. Such a method, however, would require extensive pre-training (they pre-train on internet scale data). \citet{song2025historyguidedvideodiffusion} propose attention between noisy versions of conditioning and action trajectory at training time and temporal classifier free guidance at sampling time. Their model, however, is a video prediction model, which predicts both future actions and observations and can be data and compute intensive.

\subsection{Context Length in Reinforcement Learning in Robotics}
Reinforcement learning in robotics frequently provides solutions for more dynamic tasks in robotics, and RL policies therefore often use longer context lengths than seen in imitation learning (though these observations are typically in joint space, unlike RGB observations seen in imitation learning). Some RL works in robotics have provided explicit comments on context length. \cite{li2024reinforcementlearningversatiledynamic} use a context of 66 (2 seconds at 33 Hz). Along with the entire history provided as an encoded embedding, they provide the recent 4 frames directly to the base model, arguing that full history and recent history are used differently and therefore recent history should additionally be provided through separate channels. \cite{kumar2021rmarapidmotoradaptation} split the policy into a base policy and an adaptation module. The adaptation module takes a long history and produces latent embeddings at a lower frequency, while the base policy only takes recent state and this latent embedding to produce control at a higher frequency. \cite{Peng_2018} and \cite{singh_2023} provide ablations on comparing feed-forward policies with history with LSTM based policies. \cite{Peng_2018} argue that a non-state history based LSTM better generalizes to dynamics of a physical system, while \cite{singh_2023} choose a non-history based policy as the best performer based on their ablations.

\subsection{Context Length Understanding in LLMs}
We believe the robotics community needs to build an understanding of policy learning in the presence of longer contexts. LLM community has considerable work exploring the factors that impact performance of LLMs with longer conditioning contexts. For instance, \cite{liu2023lostmiddlelanguagemodels} show that LLM performance for long-context inputs significantly depends on the location of relevant information in the input. Specifically, performance degrades if the most relevant information is in the middle of the model input history. \cite{li2024looglelongcontextlanguagemodels} create a benchmark for long-context understanding in LLMs and show that LLMs struggle with "intricate long dependency tasks". \cite{Qin_2023} show that certain attention mechanisms can lead to insufficient attention being paid to distant tokens in longer contexts.

\section{Diffusion Policy Details}
\label{appx:diffusion_policy_details}

\subsection{Overview}
Diffusion Policy is a state-of-the-art behavior cloning method introduced by \citet{diffusion_policy} which learns the observation conditional action distribution mentioned in Equation~\ref{eq:diffusion_action_distribution} by minimizing the action prediction loss
\begin{equation}
    \mathcal L_\mathcal D(\theta) = \mathbb E_{k,(\mathbf{o},\mathbf{a})\sim \mathcal D,\epsilon\sim \mathcal N(0,I)}[\lVert \epsilon - \epsilon_\theta(\alpha_k\mathbf{a} + \sigma_k\epsilon,\ \mathbf{o},\ k) \rVert^2_2]
\label{eq:denoiser_loss}
\end{equation}
where $\alpha_k$ and $\sigma_k$ are parameters related to the diffusion noise schedule ($k$ represents diffusion timestep). Essentially, the model learns to regress over noisy versions of action chunks from observation and diffusion timestep input, and at inference time uses a denoiser to conditionally sample from the action chunk distribution. 

Diffusion Policy is implemented through a combination of observation encoder and action denoising backbone. As a summary: each camera $c$ produces an image stream that comprises $T_o$ images $\{o^c_i, i\in \{t-T_o+1, \dots, t\}\}$. For each stream, an image encoder converts each of these $T_o$ images into latent embeddings $e^c_i$. \emph{Note that the same vision backbone is used for all timesteps for each camera stream}. These embeddings, concatenated with diffusion timestep embedding and robot proprioception, are passed to the downstream model for conditioning. Common choices for the downstream conditioning model are \textbf{UNet} and \textbf{DiT}. Various methods like FiLM \cite{perez2017filmvisualreasoninggeneral} and cross-attention can be applied to merge the conditioning with the denoiser. 

We choose ResNet18 \cite{he2015deepresiduallearningimage} as our image embedding backbone. Unless noted otherwise in the specific sections, the ResNet18 version from robomimic \cite{robomimic2021} is trained from scratch for our policies, with random weight initialization (similar to \citet{diffusion_policy} and \citet{wei2025empiricalanalysissimandrealcotraining}). 

\subsection{Comparison of Diffusion Policy Architectures}
\label{appx:dp_architecture_comparison}
Diffusion Policy was presented with the \textbf{UNet+FiLM} and \textbf{DiT} architectures in the original work \cite{diffusion_policy}. For single-task RGB policies with short context lengths, \textbf{UNet+FiLM} outperformed the \textbf{DiT} model for the given design choices and hyperparameters. Therefore, it became the choice for many works building on diffusion policy \cite{cadene2024lerobot, wei2025empiricalanalysissimandrealcotraining, robomimic2021, ren2024diffusionpolicypolicyoptimization}. In this work, we suggest using \textbf{UNet+Cross-Attention} as the method for training single-task long-context diffusion policies from scratch in the limited data regime. 

As an assistance to the reader, we now provide an overview of the 3 methods based on \cite{diffusion_policy, ronneberger2015unetconvolutionalnetworksbiomedical, rombach2022highresolutionimagesynthesislatent}. All methods involve passing observation inputs through an encoder that is shared across timesteps, but is often different for cameras (though it can be the same). After that, each observation embedding is concatenated with the corresponding proprioception information. The 3 architectures compared in this work differ in flow beyond this point: 
\begin{itemize}[leftmargin=1.5em, itemsep=1pt, topsep=0pt, parsep=0pt]
    \item \textbf{UNet+FiLM}: The observation embeddings from each timestep and each camera are concatenated into one long vector. A diffusion timestep embedding is further concatenated to this. The action denoising backbone is made of a UNet with intermediate 1D convolution layers. The long vector is projected to dimensions matching these layers through separate MLPs, and merged via FiLM \cite{perez2017filmvisualreasoninggeneral}, where a scale and bias are both applied. 
    \item \textbf{UNet+Cross-Attention}: The observation embeddings from each timestep are concatenated across cameras and proprioception,  but remain separate for different timesteps. Diffusion time embedding is projected to matching dimension. A position encoding is added. Each of these embeddings are then individually projected to intermediate UNet layer dimensions, and cross attended to with the intermediate UNet layers. 
    \item \textbf{DiT}: The action denoising backbone is now provided by transformer decoder layers as opposed to UNet. Conditioning is merged via cross-attention. 
\end{itemize}

We note that the \textbf{DiT} model from the original work as well as from key long-context learning work from \citet{torne2025learninglongcontextdiffusionpolicies} uses hyperparameters which result in a model that is significantly smaller in size than other choices. If we use their hyperparameters, we notice extremely low success rates in some tasks even with short context lengths. Very specifically, we notice reasonable performance on robomimic tasks. But we note almost catastrophic performance on tasks implemented in Drake \cite{drake}. We therefore add more decoder layers, providing the model number of parameters similar in magnitude to the other architectures we have used. Moreover, a key change in our experiments for \textbf{DiT} is not using causal attention within the action chunk. Both \citet{diffusion_policy} and \citet{torne2025learninglongcontextdiffusionpolicies} apply causal attention within an action chunk as it is denoised (so earlier elements of the action chunk cannot be influenced by later elements). We remove this flag, to make the flow of information more comparable to that of the \textbf{UNet}. Finally, we evaluate the \textbf{DiT} with additional steps of sampling compared to the UNet models, to eliminate the possibility of sampling error driving difference in performance. 

Finally, while we have not seen this in robotics policies, merging conditioning via cross-attention instead of FiLM has been used in computer vision before \cite{rombach2022highresolutionimagesynthesislatent}.

\subsection{Training Details and Hyperparameters}
\label{appx:diffusion_policy_training}
\subsubsection{Checkpoint Selection}
We designate a maximum number of training steps for each task based on approximate policy performance to convergence for short-context policies at $N$ training trajectories. We then train policies with all context lengths for that task to those many training steps, saving the top 5 checkpoints with respect to validation loss, top 5 checkpoints with respect to action matching validation loss, and 3 evenly spaced checkpoints toward the end of training. We then evaluate each checkpoint, and the highest success rate among checkpoints is reported as the success rate for the training run. For our hardware evaluation on the \textbf{marshmallows} task, we simply use the latest checkpoint. 

We train all \textbf{push-T} policies for 150k steps, \textbf{marshmallows}, \textbf{square}, \textbf{lift}, and \textbf{push-and-return} policies for 200k steps, and \textbf{grasp-and-return} policies for 100k steps. We start with a learning rate of \texttt{1e-4} and use \texttt{cosine} decay. For some policies (like \textbf{UNet+FiLM} and \textbf{UNet+Cross-Attention} on \textbf{push-T} and \textbf{UNet+FiLM} on \textbf{push-and-return}), we only save the latest checkpoint as opposed to 3 recent toward the end (because the results are largely insensitive to this choice, we did not rerun these configurations due to the high computational cost). We did rerun it for the $N$ data regime in \textbf{push-and-return} for \textbf{UNet+FiLM}, and found the presented trends to be consistent.  

We notice in our work that policies with different context lengths can sometimes converge in different number of training steps. For instance, we usually notice that especially with $N/2$ trajectories, some policies often converge slightly sooner than those for $N$ and $2N$, an effect that is more pronounced at longer context lengths. We believe earlier checkpoints at longer context lengths in the low data regime help mitigate the impact of over-fitting. 

Each reported success rate is from one training run. While we do account for uncertainty in the binary task success, we do not, in this work, account for uncertainty in training randomness. Due to limited availability of resources and high amount of training time needed to produce a study of this magnitude, such a choice is consistent with prior insightful work \cite{wei2025empiricalanalysissimandrealcotraining}. 

\subsubsection{Hyperparameters}
In Table~\ref{tab:arch-hparams} we present the hyperparameters used for the results in Sections~\ref{sec:performance_trends} and \ref{sec:lc_policy_learning}. By design, FiLM conditioning model size changes considerably as context length is increased. On the contrary, \textbf{UNet+Cross-Attention} and \textbf{DiT} have similar model sizes across context lengths due to the manner of applying conditioning. We choose the size of the cross-attention model to be roughly half way between the smallest and largest used FiLM model. For \textbf{DiT}, we get a model comparable in size to the \textbf{cross-attention} model after adding significantly more layers to the model from \citet{torne2025learninglongcontextdiffusionpolicies} and \citet{diffusion_policy}. 

\begin{table}[h]
\centering
\small
\caption{Architecture-specific hyperparameters for the three denoisers. Values not applicable to a given architecture are marked ``--''.}
\label{tab:arch-hparams}
\begin{tabularx}{\textwidth}{l*{3}{>{\centering\arraybackslash}X}}
\toprule
& \textbf{UNet+FiLM} & \textbf{UNet+xAttn} & \textbf{DiT} \\
\midrule
\multicolumn{4}{l}{\textit{Denoiser core}} \\
Conditioning mechanism & FiLM (global) & Cross-attention (token seq.) & Cross-attention (token seq.) \\
UNet Down channels & $[256,512,1024]$ & $[256,512,1024]$ & -- \\
Conv kernel size & 5 & 5 & -- \\
GroupNorm groups & 8 & 8 & -- \\
\midrule
\multicolumn{4}{l}{\textit{Approximate parameter counts}} \\
Denoiser parameters & $\approx 67{\rm M}$ at $h_y=2$ to $\approx 212{\rm M}$ at $h_y=80$ & $\approx 129{\rm M}$ & $\approx 109{\rm M}$ \\
Encoder parameters & $22.4{\rm M}$ & $22.4{\rm M}$ & $22.4{\rm M}$ \\
Total parameters & $\approx 89{\rm M}$ at $h_y=2$ to $\approx 235{\rm M}$ at $h_y=80$ & $151{\rm M}$ & $131{\rm M}$ \\
\bottomrule
\end{tabularx}
\end{table}

Task specific hyperparameters are available in Table~\ref{tab:task-hparams}. Each task differs in image resolution, crop size, action dimensionality, the set of observation horizons we sweep, the number of training demonstrations used, and the total number of training steps. Total training steps are matched between baselines and ablations to ensure a fair comparison. 

Our models are designed such that the time dimension of the trajectory being denoised needs to be a multiple of 4. We select a total denoising horizon such that the length of the predicted future is roughly 16 while retaining full past action prediction and having the total length a multiple of 4. Thus, for context length 1, our horizon is 16, for 2, it is also 16, for context length 5 it is 20, and for context length 10 it is 24. If the context length is a multiple of 4, the horizon is simply context length + 16. Further details are in Table \ref{tab:task-hparams}.

While this is an accurate tabular representation, the policy design involves many more hyperparameters which cannot be best captured in this manner. We therefore refer the reader to our website and codebase for the updated and most accurate configuration files corresponding to our experiments. 

\begin{table}[h]
\centering
\scriptsize
\setlength{\tabcolsep}{3pt}
\caption{Some details specific to the task evaluated.}
\label{tab:task-hparams}
\begin{tabularx}{\textwidth}{l*{6}{>{\centering\arraybackslash}X}}
\toprule
\textbf{Setting} & \textbf{Push-T} & \textbf{Push-and-} & \textbf{Grasp-and-} & \textbf{Marshmallows} & \textbf{Square} & \textbf{Lift} \\
 & & \textbf{return} & \textbf{return} & \textbf{(hardware)} & \textbf{(robomimic)} & \textbf{(robomimic)} \\
\midrule
Image input ($H{\times}W$) & $128{\times}128$ & $128{\times}128$ & $128{\times}128$ & $128{\times}128$ & $84{\times}84$ & $84{\times}84$ \\
Crop ($H{\times}W$) & $112{\times}112$ & $112{\times}112$ & $112{\times}112$ & $112{\times}112$ & $76{\times}76$ & $76{\times}76$ \\
Cameras & overhead, wrist & overhead, wrist & overhead, wrist & overhead, wrist & agentview, eye-in-hand & agentview, eye-in-hand \\
Proprioception dim & 3 & 2 & 13 & 13 & 9 & 9 \\
Action dim & 2 & 2 & 13 & 13 & 10 & 10 \\
Context length swept ($h_y$) & 1, 2, 5, 10, 12, 16, 20 & 4, 8, 32, 60, 72, 80 & 4, 8, 16, 20, 24, 48 & 92 & 1, 2, 5, 10, 12, 16, 20 & 1, 2, 5, 10, 12, 16, 20 \\
Future chunk length $h_u$ & $h_y$+14 to $h_y$+16 & $h_y + 16$ & $h_y + 16$ & $h_y + 16$ & $h_y + 14$ to $h_y + 16$ & $h_y + 14$ to $h_y + 16$ \\
Data (low\,/\,usual\,/\,high) & 80\,/\,160\,/\,320 & 24\,/\,48\,/\,96 & 12\,/\,24\,/\,48 & 100 & 50\,/\,100\,/\,200 & 25\,/\,50\,/\,100 \\
Total training steps & $1.5{\times}10^{5}$ & $2{\times}10^{5}$ & $1{\times}10^{5}$ & $2{\times}10^{5}$ & $2{\times}10^{5}$ & $2{\times}10^{5}$ \\
\bottomrule
\end{tabularx}
\end{table}

\section{Task Details}
\label{appx:task_details}
The value of $N$ used for each task is indicated in Table~\ref{tab:data_per_task}. We then present further details about data collection (Section~\ref{appx:data_collection}) and evaluation (Section~\ref{appx:evaluation_details}) for all the tasks. 

\begin{table}[htbp]
    \centering
    \caption{Number of training trajectories chosen as $N$ for different tasks used in Section~\ref{sec:performance_trends}. The values for benchmark tasks are chosen to be consistent with prior work \cite{wei2025empiricalanalysissimandrealcotraining, robomimic2021}.}
    \begin{tabular}{lc}
        \toprule
        \textbf{Task} & \textbf{Value of} $\mathbf{N}$ \\
        \midrule
        \textbf{push-T} & 160 \\
        \textbf{robomimic square} & 100 \\
        \textbf{robomimic lift} & 50 \\
        \textbf{push-and-return} & 48 \\
        \textbf{grasp-and-return} & 24 \\
        \textbf{marshmallows} & 100 \\
        \bottomrule
    \end{tabular}
    \label{tab:data_per_task}
\end{table}

\subsection{Task and Data Collection Descriptions}
\label{appx:data_collection}
\textbf{Push-T:} Observations are images \textit{(from pixels)} and robot proprioception, whereas actions are a chunk of the next planar location of the cylindrical end effector. Dataset for this task was collected using teleoperation and is available via \cite{wei2025empiricalanalysissimandrealcotraining}. We choose $N=160$.

\textbf{Robomimic Tasks:} We use the proficient human dataset. Further details on the environments are available via \textbf{robomimic} \cite{robomimic2021}. We choose $N=100$ for \textbf{square} and $N=50$ for \textbf{lift}. 

\begin{figure}[!tbp]
    \centering

    \begin{subfigure}[t]{0.31\textwidth}
        \centering
        \includegraphics[height=3.8cm]{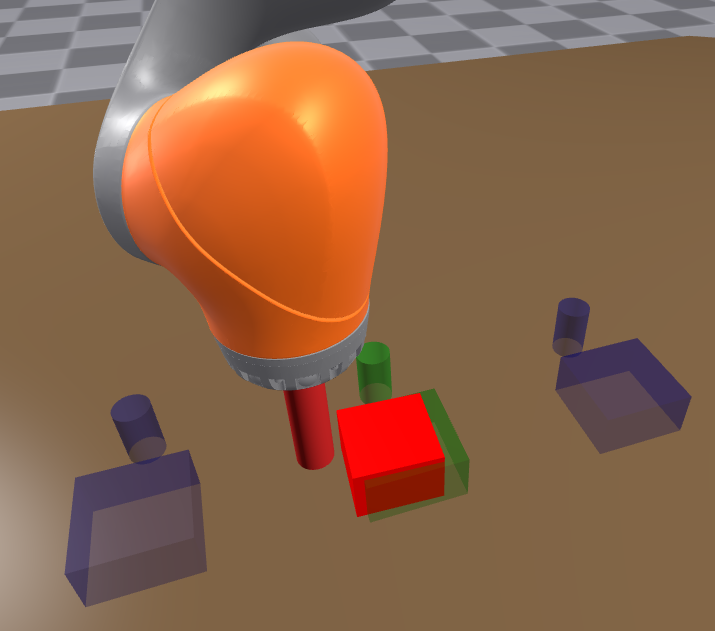}
        \caption{}
        \label{fig:task_images_push_return}
    \end{subfigure}
    \hfill
    \begin{subfigure}[t]{0.31\textwidth}
        \centering
        \includegraphics[height=3.8cm]{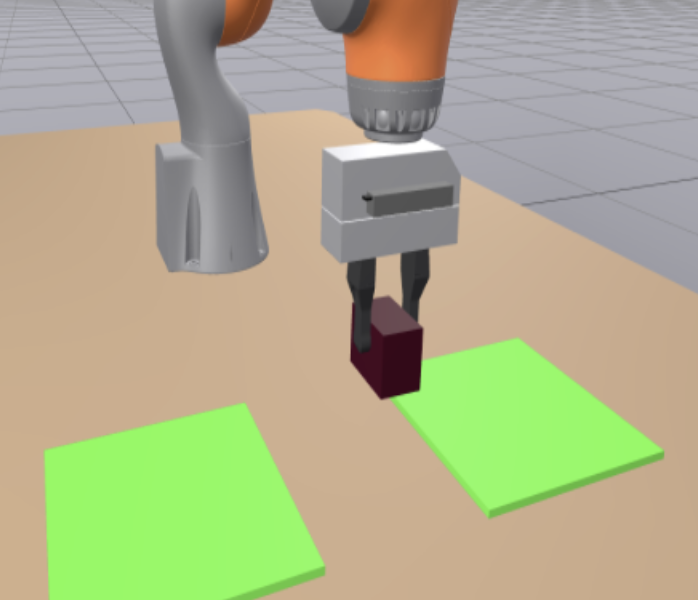}
        \caption{}
        \label{fig:task_images_grasp_return}
    \end{subfigure}
    \hfill
    \begin{subfigure}[t]{0.31\textwidth}
        \centering
        \includegraphics[height=3.8cm]{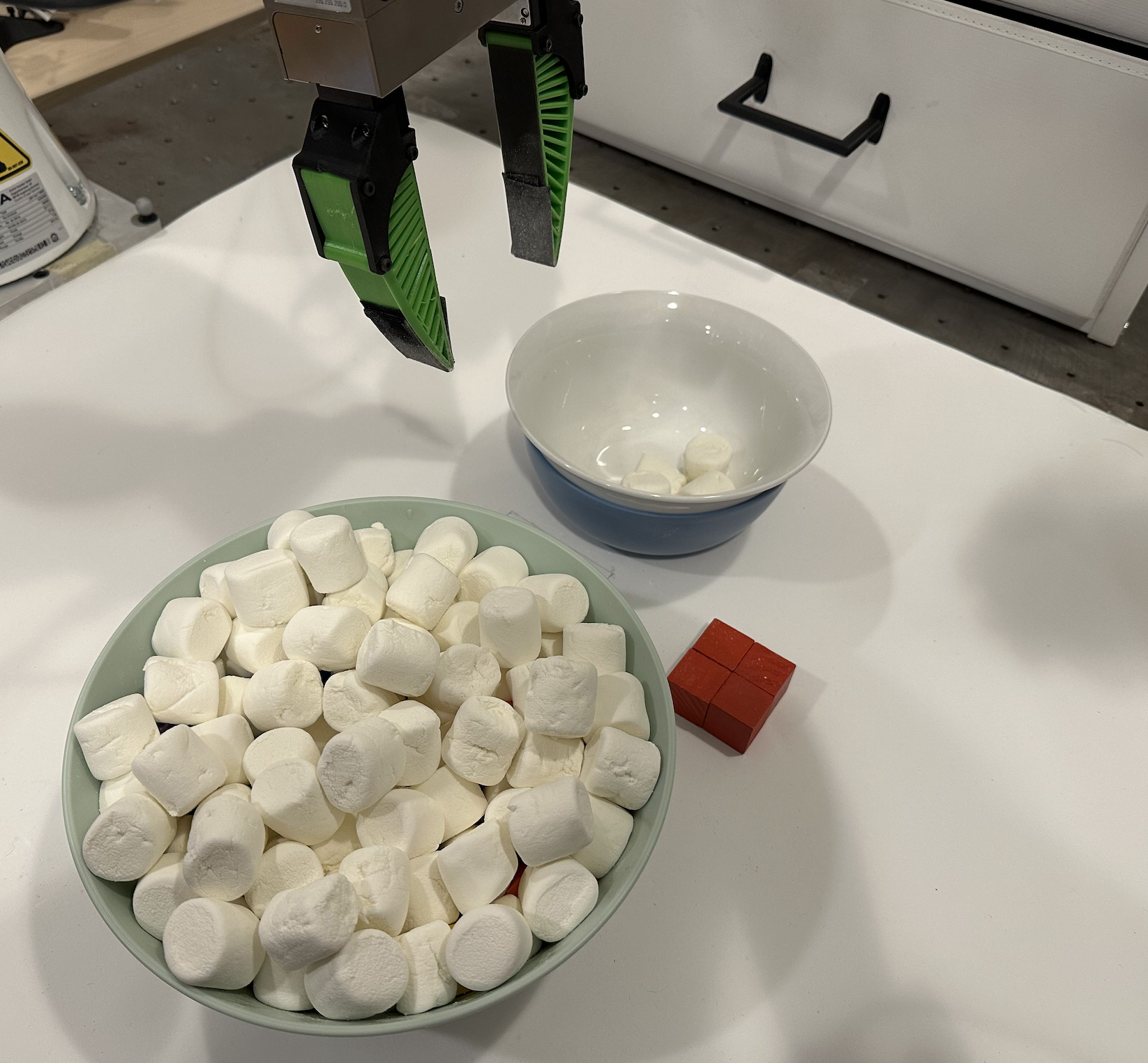}
        \caption{}
        \label{fig:task_images_hardware_setup_marshmallows}
    \end{subfigure}

    \caption{Environments for tasks requiring long context length for success. \textbf{(a)} Push-and-return \textbf{(b)} Grasp-and-return \textbf{(c)} Marshmallows.}
    \label{fig:task_setups}
\end{figure}

\textbf{Push-and-Return:} Observation and action space are the same as defined above for \textbf{push-T}. We collect data for this task via teleoperation in simulation. For each mode, we collect two datasets: (i) by randomly initializing the block in that location and completing the task via teleoperation and (ii) by mirroring tele-operated roll-outs from the symmetrically opposite mode. This allows us to ensure coherent data across symmetric modes, and eliminate inconsistent teleoperation as a source of learning discrepancy. We use a two-mode version for all the demonstrated comparisons and leave it to future work to evaluate the same task with more return modes. Such a future experiment can provide insight on increasing combinatorial complexity while keeping local complexity the same. We choose $N=48$. See Figure~\ref{fig:task_images_push_return} for reference. 

\textbf{Grasp-and-Return:} Observations remain the same as in push-and-return, but action is now 6-DOF end-effector pose as opposed to planar location as in previous cases. Data for this task is generated in simulation using heuristic method adapted from \cite{manipulation}. We choose $N=24$. See Figure~\ref{fig:task_images_grasp_return} for reference. 

\textbf{Marshmallows: }This task requires moving two handfuls of marshmallows to a target bowl from a supply bowl, and then signaling task completion by pressing a button (we represent this by gripper closing and reaching a small platform of red blocks). We adapt this task on our hardware from \cite{mark2026bpplongcontextrobotimitation}, who in turn have adapted it from a task in \cite{torne2025learninglongcontextdiffusionpolicies}. We collect data using teleoperation via VR. Observations are provided via one overhead camera and one wrist camera. Action space is the same as in \textbf{grasp-and-return}. We choose $N=100$. See Figure~\ref{fig:task_images_hardware_setup_marshmallows} for reference. 

\subsection{Evaluation Details}
\label{appx:evaluation_details}
Reported success rates are out of 200 trials for simulated tasks with error bars representing 95\% Wilson score confidence intervals.  

\textbf{Push-T:} Specific evaluation details for this task are based on \cite{wei2025empiricalanalysissimandrealcotraining}. 

\textbf{Push-and-return:} In the 2-mode case, we conduct 100 trials with the block originating on either side. The time limit is set to 50s. The block is considered moved to the middle and back to either location if it overlaps by more than 95\% of its area with the marker for that location.

\textbf{Grasp-and-return:} We conduct 100 trials with the brick originating on either side. The brick is considered moved to the center if it makes contact with the table space between the mats on either side. The brick is considered returned to a location if it is returned to anywhere on the mat from which it originated. 

\textbf{Marshmallows:} We conduct 20 trials on hardware. The number of marshmallows at start of the task in the target bowl can be arbitrary (consistent with data collection, and makes the task challenging in terms of memory). 

\section{Additional Discussion}
\label{appx:additional_discussion}

\subsection{Reasons for Prior Works' Criticism of Naive History Scaling}
\label{appx:naive_scaling_criticism}
In Section~\ref{subsec:empirical_validation_performance_trends}, we show that extreme failure as indicated in some prior works is not noted with naive context length scaling. However, some prior works \cite{torne2025learninglongcontextdiffusionpolicies, sridhar2025memerscalingmemoryrobot} do advertise that naively scaling context length can lead to stark failure. We elaborate on our explanation for why perhaps they do so: 
\begin{enumerate}
    \item \textbf{Not benchmarking on dataset size: }As shown through results in Figures \ref{fig:push_and_return_cross_task_success_data_scaling}, \ref{fig:grasp_and_return_cross_task_success_data_scaling}, \ref{fig:push_t_data_cl_scaling}, \ref{fig:square_data_cl_scaling}, and \ref{fig:lift_data_cl_scaling}, increasing data leads to significant gains at longer context lengths and the relative difference between short and long context policies changes. Notably, tasks show different sample complexity at longer context lengths (locally stable grasping tasks like \textbf{lift} and \textbf{grasp-and-return} show success even in the case of $N/2$ trajectories at longer context lengths). Thus, without benchmarking on dataset size, it is unclear if the policy is being tested in the low data regime for the task or if naive scaling is algorithmically catastrophic. Data scale investigation from \citet{mark2026bpplongcontextrobotimitation}, while not as detailed as our work, does support our claims. 
    \item \textbf{Inappropriate architecture choice: }As we show in section \ref{sec:cross_attention_conditioning}, the choice of conditioning method significantly impacts long-context policy learning. Vast majority of diffusion policy works use \textbf{UNet + FiLM} \cite{cadene2024lerobot, wei2025empiricalanalysissimandrealcotraining, robomimic2021, ren2024diffusionpolicypolicyoptimization}. Some other works \cite{torne2025learninglongcontextdiffusionpolicies} use \textbf{DiT} implementation from \citet{diffusion_policy}, which is indicated to be a poorly performing denoising backbone for single-task policies from image inputs \cite{diffusion_policy} and should not be used for benchmarking single task policies. We show in Section~\ref{sec:cross_attention_conditioning} that even with favorable hyperparameter changes, the DiT fails to perform in the simulation tasks we define in Drake \cite{drake}. 
    \item \textbf{Lack of Task Comparisons: }Our results comparing Figures \ref{fig:push_and_return_cross_task_success_data_scaling} and \ref{fig:push_t_data_cl_scaling} show that even if some methods perform well on tasks requiring memory to succeed, they may still show diminishing returns on tasks solvable with short histories. As we incrementally build solutions toward general purpose policies, it is necessary to provide a long-context learning solution which performs well across the spectrum of tasks needing and not needing long memory to solve. Therefore, it is necessary to benchmark solutions on tasks solvable with short history when proposing long-context learning solutions. 
\end{enumerate}

\subsection{Impact of Grasp as Manipulation Primitive}
\label{appx:grasp_as_manipulation_primitive}
In Section~\ref{sec:performance_trends}, we show that tasks like \textbf{lift} and \textbf{grasp-and-return} show reasonable long-context performance even in the low data regime ($N/2$). We speculate that this is because these tasks involve locally stable manipulation primitives, and policies can perform well despite errors in manipulation. 

\begin{figure}[!tbp]
    \centering
    \includegraphics[width=0.3\linewidth]{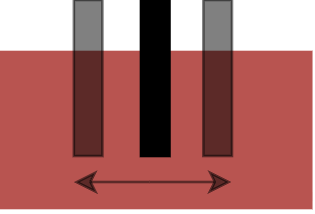}
    \caption{During data collection, the robot grippers make contact with the brick along its central axis while grasping. At inference time however, the policy commands grippers to make contact away from the center location, indicating error in learning the manipulation skills from data. However, because this task is locally stable, despite this error policies perform well. This task thus shows lower sample complexity for long-context learning.}
    \label{fig:task_3_brick_position}
\end{figure}

To support our claim, we study \emph{how well did a long-context policy learn to grasp} for the \textbf{grasp-and-return} task in the $N$ data regime. We examine the grasping skill learned in \textbf{grasp-and-return} by comparing the location where gripper makes contact with the brick relative to the center of the brick. As shown in Figure \ref{fig:task_3_brick_position}, training data has gripper making contact at the center of the brick (we can control for this as the data is synthetically generated with a heuristic method). We show the mean and median absolute value of gripper contact location with respect to center of the brick at inference time (policy rollout) in Table~\ref{tab:contact_offsets}. Despite having errors more than 1 cm, the task shows success at long contexts. This shows that for tasks where it is possible to succeed even with significant errors in the learned manipulation skill, long-context policies are able to learn even at low sample complexity. 

\begin{table}[htbp]
\centering
\caption{Mean and median of the absolute value of gripper contact location offset from the center of the brick in meters. The value for expert truth data is 0. Despite most values being more than 1 cm away from the center, long-context policies succeed ($N$ training trajectories).}
\setlength{\tabcolsep}{4pt}
\renewcommand{\arraystretch}{1.1}

\begin{tabular}{lcccccc}
\toprule
\textbf{Context Length} 
& 20 
& 24 
& 48 \\
\midrule

\textbf{Mean}
& 0.012
& 0.014
& 0.013
\\

\textbf{Median}
& 0.010
& 0.010
& 0.010
\\

\bottomrule
\end{tabular}
\label{tab:contact_offsets}
\end{table}

\subsection{Reasons for Long-Context Failure}
\label{appx:reasons_for_long_context_failure}
In Section~\ref{sec:performance_trends} we saw how long-context policies do not show catastrophic failure in most cases. However, Figures \ref{fig:push_t_data_cl_scaling}, \ref{fig:square_data_cl_scaling}, and \ref{fig:push_and_return_cross_task_success_data_scaling} still show that especially in the low data regime ($N/2$ trajectories), long-context policies still perform worse than short-context ones for some tasks. In this section, we further diagnose the reasons behind these behaviors.

\subsubsection{Training and test loss}
Section \ref{appx:diffusion_policy_training} explains how checkpoints are saved during training. All saved checkpoints are evaluated and the success rate of the best checkpoint is reported. We note that \textit{because the work presented in this section has only been done for one training run, the results can be noisy}.

\begin{figure}[!tbp]
\centering
\subfloat[]{%
    \includegraphics[width=0.48\textwidth]{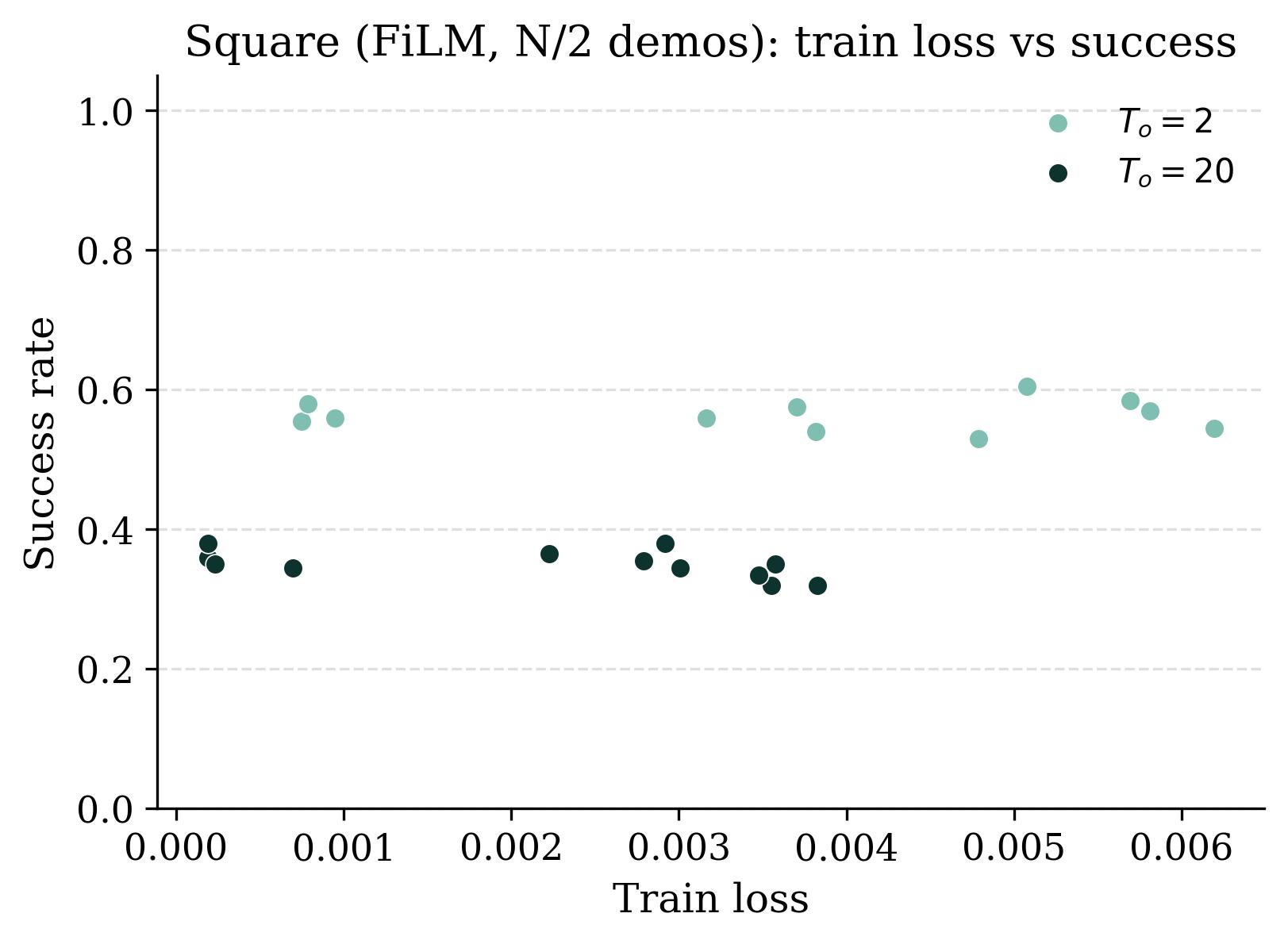}
    \label{fig:scratch_loss_vs_success_data50_train}
}
\hfil
\subfloat[]{%
    \includegraphics[width=0.48\textwidth]{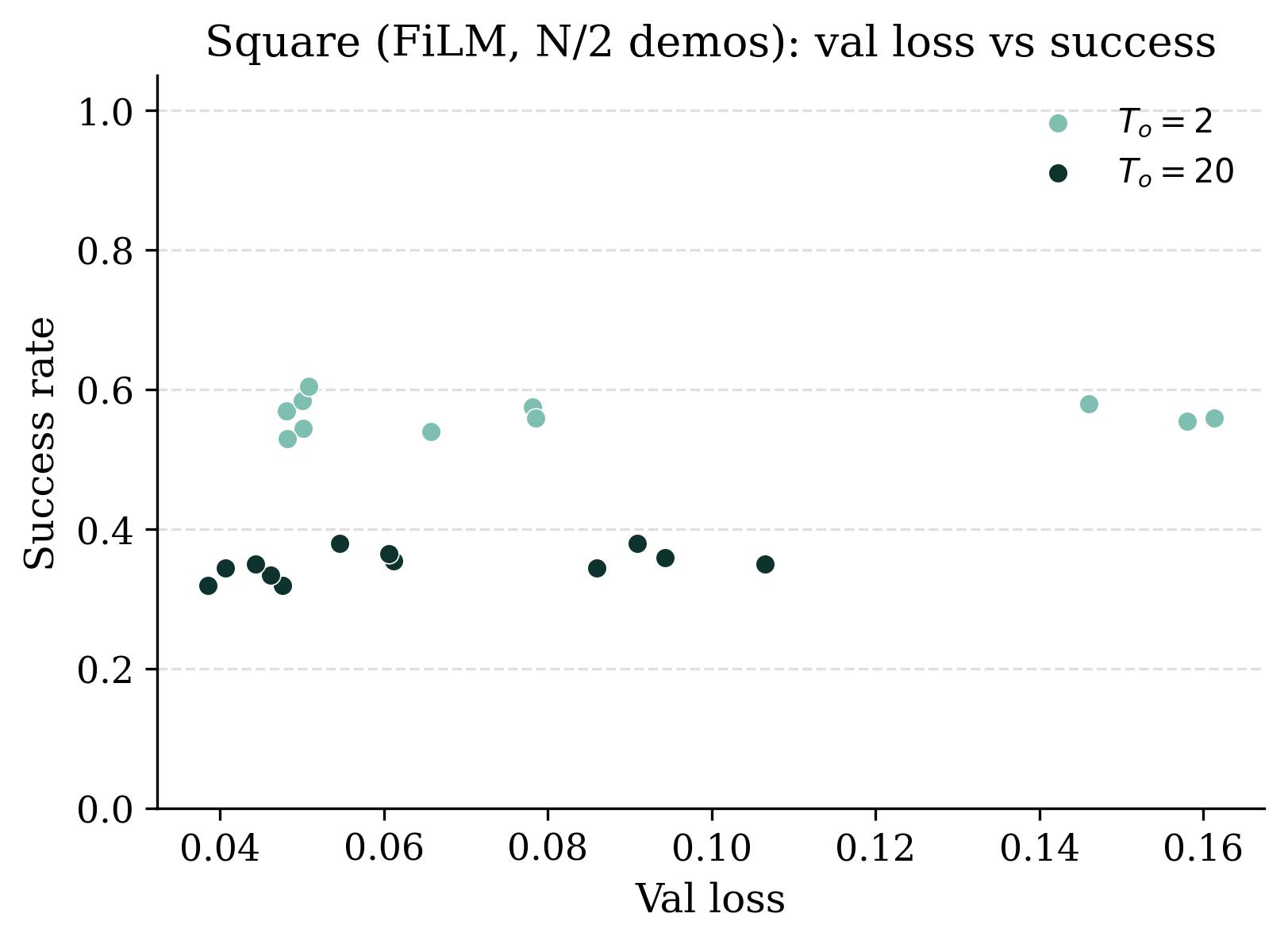}
    \label{fig:scratch_loss_vs_success_data50_val}
}
\caption{\textbf{Relationship between loss and success rate for the square task across checkpoints.}
(a) Training loss vs.\ success rate.
(b) Validation loss vs.\ success rate.}
\label{fig:loss_vs_success_square}
\end{figure}

In Section \ref{sec:cross_attention_conditioning}, we show that some architectures like \textbf{UNet+FiLM} show a sharper drop as context length is increased as compared to \textbf{UNet+Cross-Attention}. For \textbf{UNet+FiLM}, consider Figures \ref{fig:scratch_loss_vs_success_data50_train} and \ref{fig:scratch_loss_vs_success_data50_val}, which compare the training and validation loss respectively against the success rate for all checkpoints saved during training for \textbf{square} task trained with $N/2$ trajectories. As context length increases from short to long, checkpoint clusters shift downward and leftward: both training and validation loss decrease overall, but so does the success rate. The trend with training loss shows a standard over-fitting problem: policies fit to the training data better in higher dimensions, but this is not sufficient for closed loop rollouts. The validation loss curve hints toward an even more interesting trend seen in imitation learning: long-context policies perform better on open-loop validation samples drawn from expert distribution, but perform worse when rolled out closed loop. This hints toward \emph{covariate shift} in imitation learning \cite{ross2011reductionimitationlearningstructured}. Both these trends hint at data scaling and relevant inductive biases as favorable solutions for long-context learning. 

As mentioned above, we show results from one training run, so results can be noisy. The best way to reproduce this experiment would thus be to have multiple training runs, and then average the clusters from checkpoints saved during each training to produce a less noisy version of Figure \ref{fig:loss_vs_success_square}. This, we believe, would also give more consistent results across architectures, checkpoints, and context lengths and is a very relevant direction for future work. 

\subsubsection{Manipulation Skill Learning}
In Section~\ref{sec:performance_trends}, we note that in the low data regime, \textbf{grasp-and-return} provides good long-context performance, where as \textbf{push-and-return} performance drops. We now comment in detail on additional evaluation criterion to investigate \emph{what do policies fail to learn as context length increases?}

For \textbf{push-and-return}, both \textbf{task success} (Figure~\ref{fig:push_and_return_cross_task_success_data_scaling}) and \textbf{manipulation completion} (Figure~\ref{fig:push_and_return_cross_subtask_success_data_scaling}) show dropping performance with increasing context length in the $N/2$ data regime. However, investigating Figure~\ref{fig:push_and_return_cross_contextual_success_data_scaling} shows that \textbf{contextual success} is high for long context policies even when trained on $N/2$ trajectories. This indicates that in the low data regime, most long-context policies are failing to learn to manipulate within the constraints of the task. In fact, those policies which do learn to complete manipulation also show good tracking of history. Thus, long-context policies for this task fail due to difficulty learning the manipulation skill within the realms of the task. Simultaneous and mirroring improvement seen in Figures~\ref{fig:push_and_return_cross_task_success_data_scaling} and \ref{fig:push_and_return_cross_subtask_success_data_scaling} confirms this hypothesis. 

We note that this simultaneous improvement between \textbf{task success} and \textbf{manipulation completion} is also noted in the \textbf{grasp-and-return} task (Figures~\ref{fig:grasp_and_return_cross_task_success_data_scaling} and \ref{fig:grasp_and_return_cross_subtask_success_data_scaling} respectively). However, both are already good for this task in the low data regime, perhaps because this task builds on locally stable manipulation skill. In Section~\ref{appx:grasp_as_manipulation_primitive} we have further commented on this aspect. 

\subsection{When are short-context lengths sufficient?}
\label{appx:when_is_short_context_sufficient}
We comment on the difference with respect to \textbf{manipulation completion} between short-context ($T_o \in \{4,8\}$) policy performance on \textbf{push-and-return} (Figure~\ref{fig:push_and_return_cross_subtask_success_data_scaling}) v/s \textbf{grasp-and-return} (Figure~\ref{fig:grasp_and_return_cross_subtask_success_data_scaling}). As expected, we notice good manipulation completion for short-context policies on \textbf{push-and-return}. But why do they perform poorly on \textbf{grasp-and-return} with respect to this metric? We qualitatively examine the rollouts and find that high-level decision-making failures caused by short context lengths can cause low-level control failures.

To illustrate this, consider a commonly observed failure case for \textbf{grasp-and-return} with $T_o=4$. The policy does not have enough memory to commit to a subroutine (ex. moving to the center or returning to the neutral position). This mode-switching produces jittery actions that drive the robot into out-of-distribution states where its manipulation abilities deteriorate. This results in unwarranted collisions, failed grasps, and ultimately unsuccessful rollouts.

This pattern reinforces the notion that there are fundamental limitations to imitating long-memory experts with short-context policies. For instance, without sufficient history, the policy may lack the information to infer the latent state underlying the expert’s long-horizon behavior. Even more interestingly, the observability of this latent state can be dependent on the manipulation primitive used - \textbf{push-and-return}, due to its planar nature, has pusher on different sides of the block. As a result, it is able to infer direction of motion and even with short context length, achieve high \textbf{manipulation completion}. \textbf{Grasp-and-return}, on the other hand, cannot distinguish phase of motion because regardless of direction of motion, grasping a block is visually similar. Consequently, overcoming these failures likely requires algorithmic changes that explicitly address long-context reasoning, rather than data scaling alone. These findings motivate approaches that either extend temporal context or provide explicit sub-task guidance (e.g., via language \cite{intelligence2025pi05visionlanguageactionmodelopenworld}).

\section{Additional Experiments}
\label{appx:additional_experiments}

\subsection{Cross-Attention Conditioning}
\label{appx:cross_atten_conditioning_additional}
\begin{figure}[!tbp]
    \centering
    \includegraphics[width=0.8\textwidth]{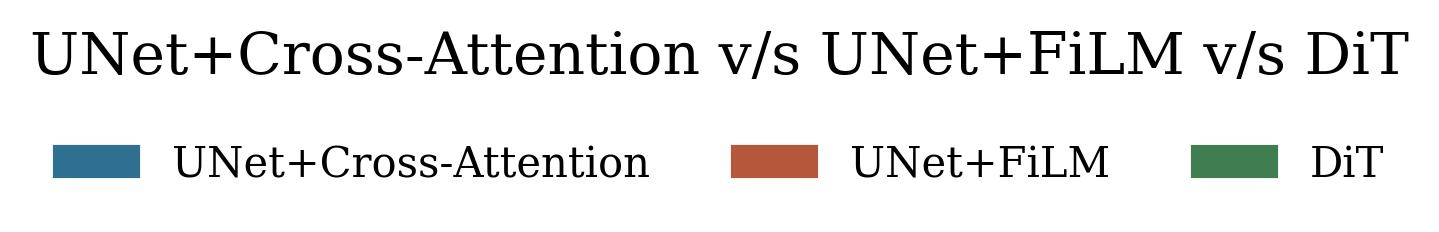}

    \medskip
    \begin{subfigure}[t]{0.47\textwidth}
        \centering
        \includegraphics[width=\textwidth]{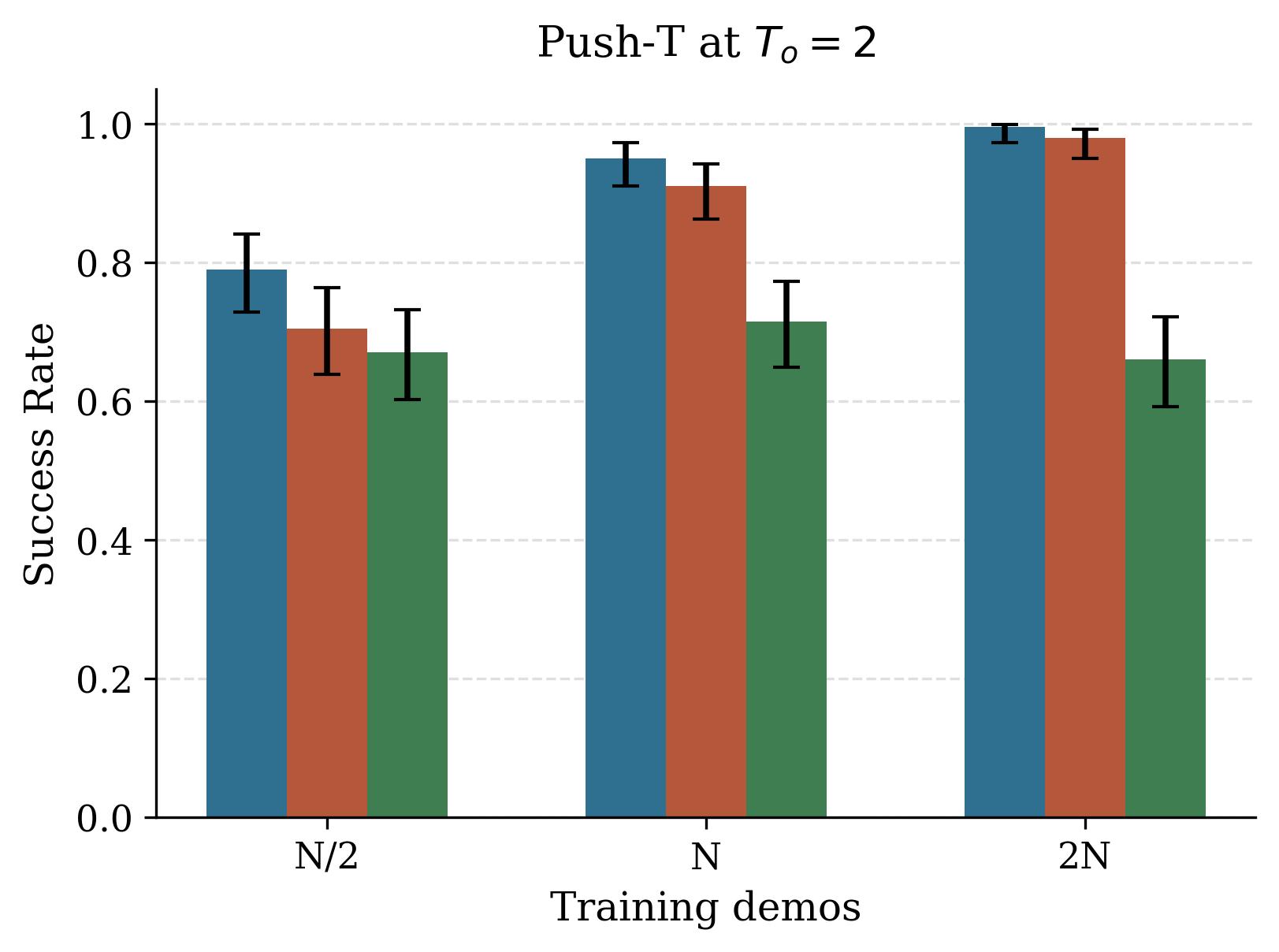}
        \caption{}
        \label{fig:push_t_unet_cross_vs_film_short}
    \end{subfigure}
    \hfill
    \begin{subfigure}[t]{0.47\textwidth}
        \centering
        \includegraphics[width=\textwidth]{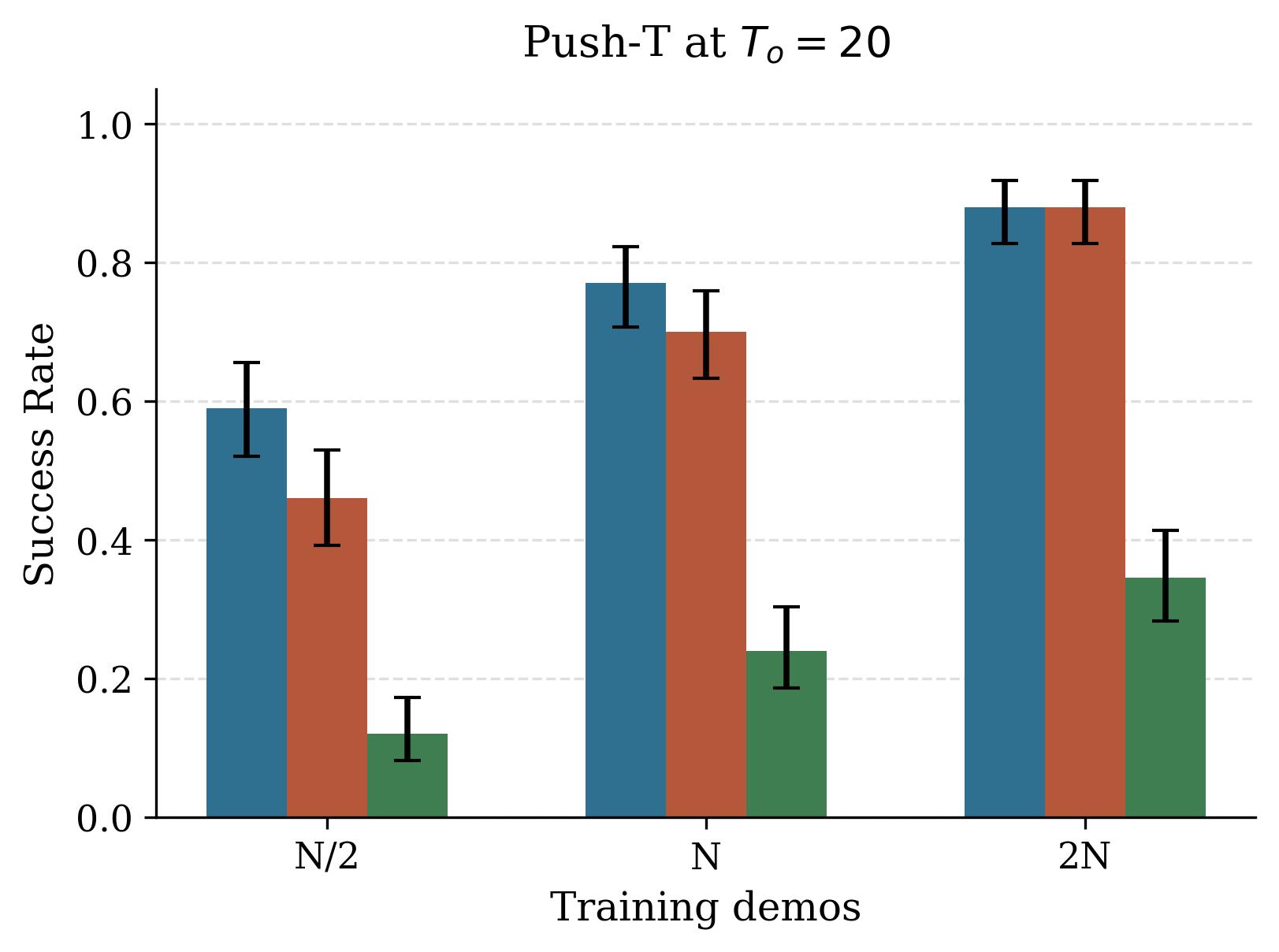}
        \caption{}
        \label{fig:push_t_unet_cross_vs_film_long}
    \end{subfigure}

    \medskip

    \begin{subfigure}[t]{0.47\textwidth}
        \centering
        \includegraphics[width=\textwidth]{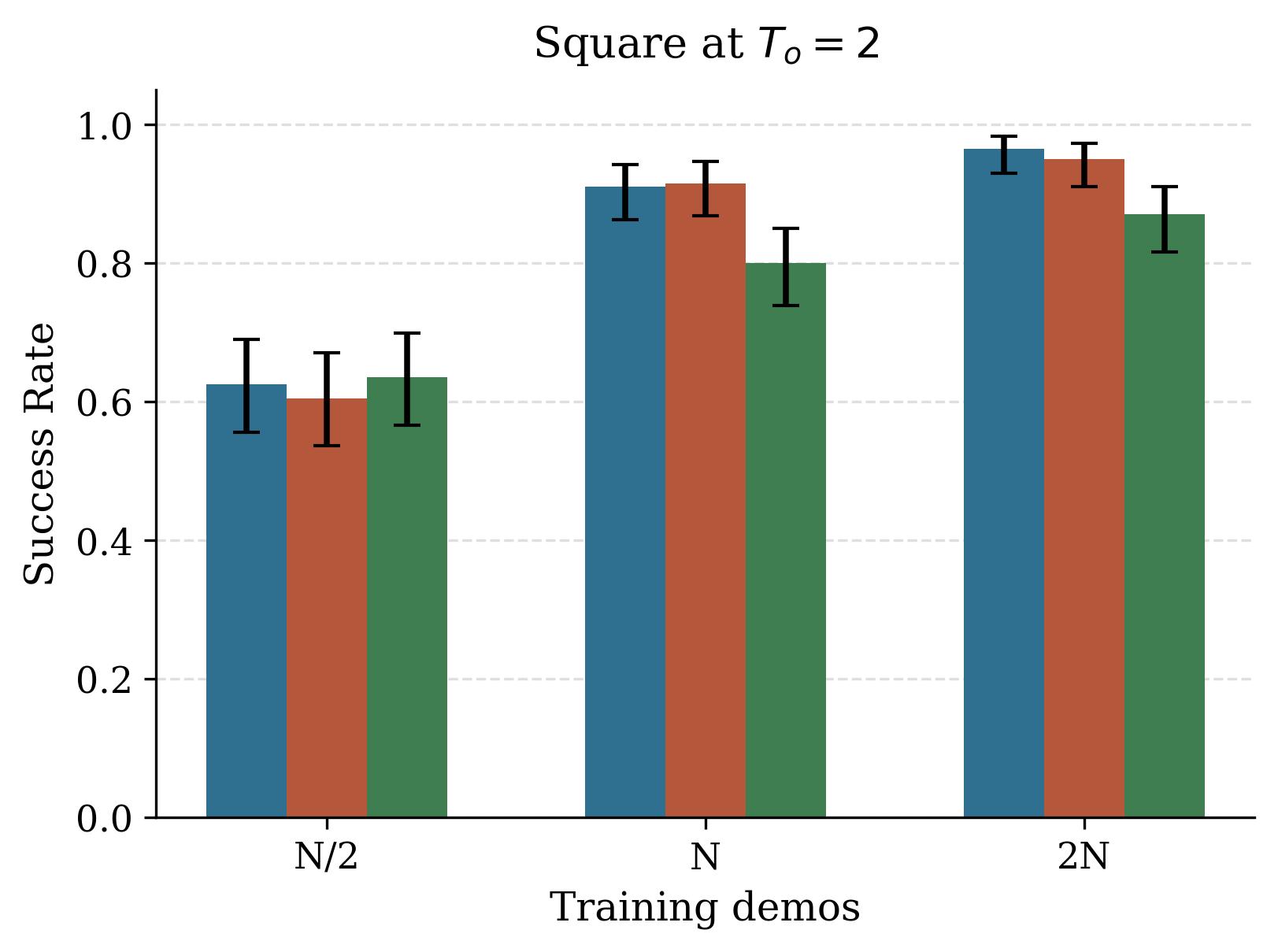}
        \caption{}
        \label{fig:square_unet_cross_vs_film_short}
    \end{subfigure}
    \hfill
    \begin{subfigure}[t]{0.47\textwidth}
        \centering
        \includegraphics[width=\textwidth]{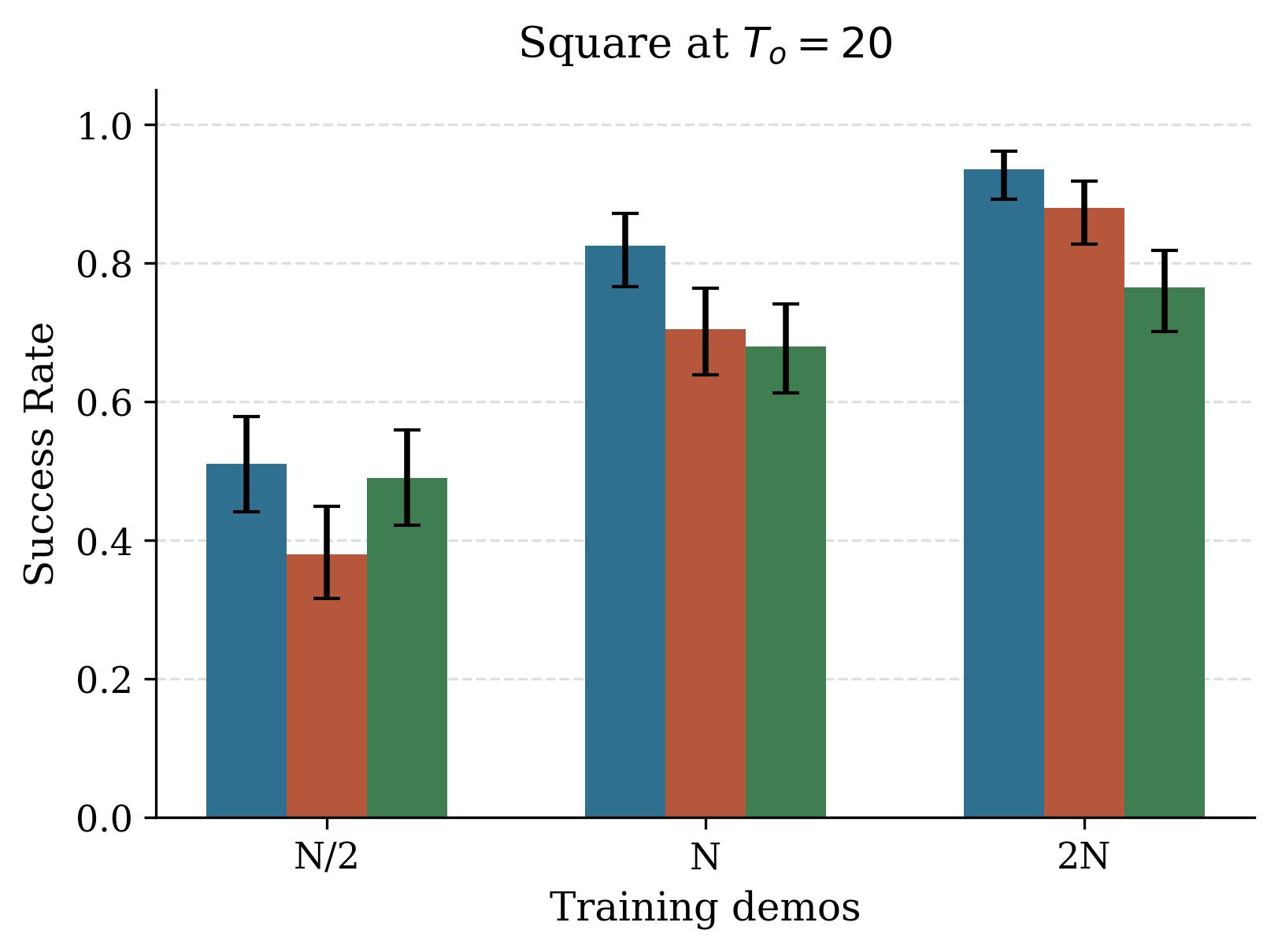}
        \caption{}
        \label{fig:square_unet_cross_vs_film_long}
    \end{subfigure}

    \medskip

    \begin{subfigure}[t]{0.47\textwidth}
        \centering
        \includegraphics[width=\textwidth]{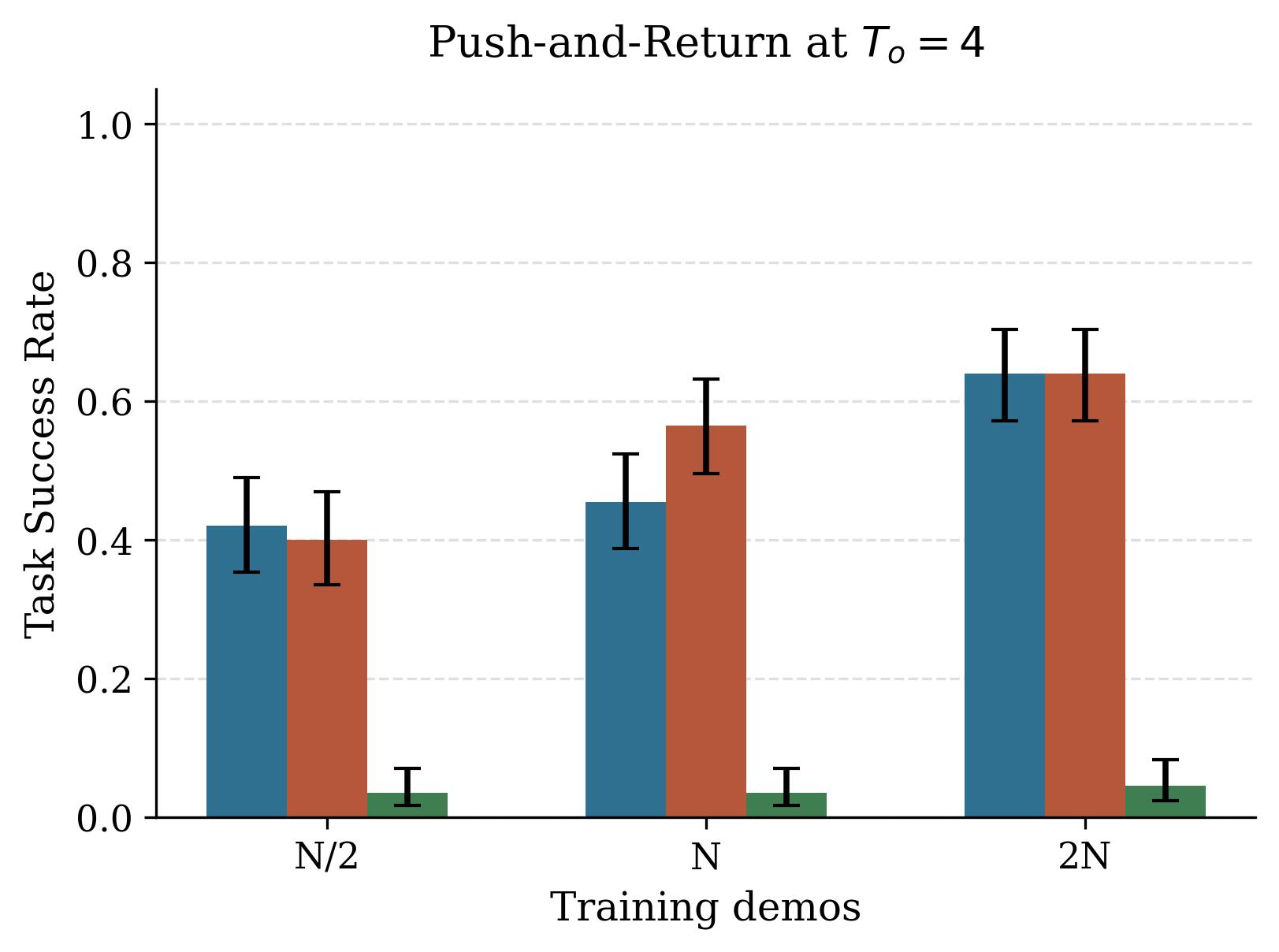}
        \caption{}
        \label{fig:push_and_return_unet_cross_vs_film_short}
    \end{subfigure}
    \hfill
    \begin{subfigure}[t]{0.47\textwidth}
        \centering
        \includegraphics[width=\textwidth]{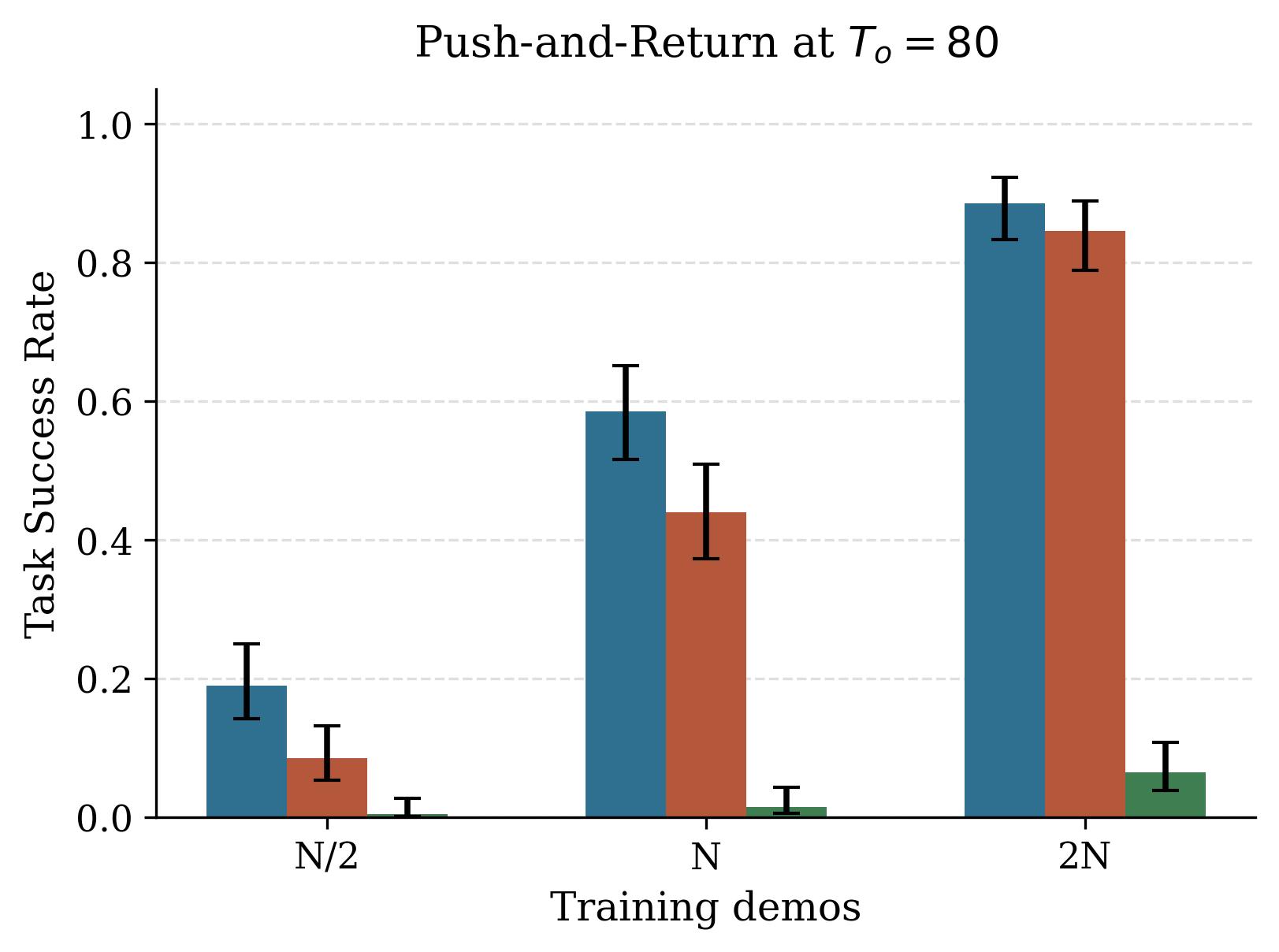}
        \caption{}
        \label{fig:push_and_return_unet_cross_vs_film_long}
    \end{subfigure}

    \caption{Comparison of \textbf{UNet + Cross Attention} and \textbf{UNet + FiLM} across tasks, context lengths, and data scales.
    Top row: \textbf{push-T} at $T_o = 2$ and $T_o = 20$.
    Middle row: \textbf{square} at $T_o = 2$ and $T_o = 20$.
    Bottom row: \textbf{push-and-return} at $T_o = 4$ and $T_o = 80$.
    Across tasks, the two architectures perform similarly at short context lengths, while \textbf{UNet + Cross Attention} outperforms \textbf{UNet + FiLM} at longer context lengths in limited-data regimes. As data is scaled, the performance gap decreases, suggesting that cross-attention provides an inductive bias that reduces the sample complexity of long-context policy learning. \textbf{DiT} is also shown as a comparison, but often fails to perform without significant pre-training in the single task cases.}
    \label{fig:unet_cross_vs_film_all_tasks}
\end{figure}

In Section~\ref{sec:cross_attention_conditioning} we establish the importance of architectural choice when comparing diffusion policy performance, especially at longer context lengths. Here we present additional results organized by different data scales at a given context length in  Figure~\ref{fig:unet_cross_vs_film_all_tasks}. While \textbf{DiT} shows no noticeable performance gain even in the high data regime, the performance gap between \textbf{UNet+Cross-Attention} and \textbf{UNet+FiLM} shrinks as data is scaled. This reinforces the notion that the cross-attention conditioning method is a favorable inductive bias. 

Additionally, Table~\ref{tab:arch-hparams} shows that the FiLM model size can be considerably different if context length is changed, while \textbf{UNet+Cross-Attention} policy is consistently at around 150M parameters. Thus, it could be particularly concerning that one comparison we have made is at context length 80, where we have compared \textbf{UNet+FiLM} with about 235M parameters with \textbf{UNet+Cross-Attention} at 151M parameters. 

\begin{figure}[!tbp]
    \centering
    \includegraphics[width=1.0\linewidth]{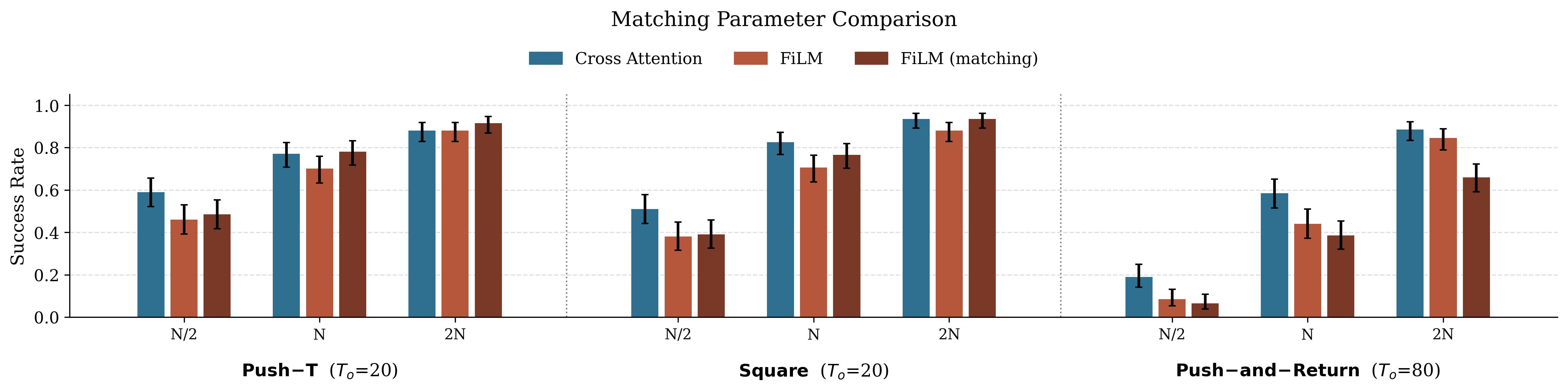}
    \caption{Comparing policy architectures when total parameters for policy using FiLM conditioning are altered to match total parameters for the cross-attention policy. }
    \label{fig:matching_parameter_comparison}
\end{figure}

The comparison we have presented in Figure~\ref{fig:unet_cross_vs_film_all_tasks} is such that the parameters in the intermediate layers of the UNet are kept consistent between \textbf{FiLM} and \textbf{cross-attention} conditioning. That is, we study what conditioning method is the best for a given UNet of certain hyperparameters. However, we now present a second comparison where we alter the size of the UNet layers to keep the total number of parameters consistent with \textbf{UNet+Cross-Attention} at a given context length. We present results in Figure~\ref{fig:matching_parameter_comparison}. Note that we have added additional parameters for \textbf{push-T} and \textbf{square} at $T_o=20$, and removed parameters for \textbf{push-and-return} at $T_o=80$, to bring the total parameters similar to those of cross-attention conditioning. We find no changes across tasks in the $N/2$ data regime.  

\subsection{Variable History Training Ablations}
\label{appx:vht_ablations}
\begin{figure}[hptb]
    \centering
    \begin{subfigure}[t]{0.47\textwidth}
        \centering
        \includegraphics[width=\textwidth]{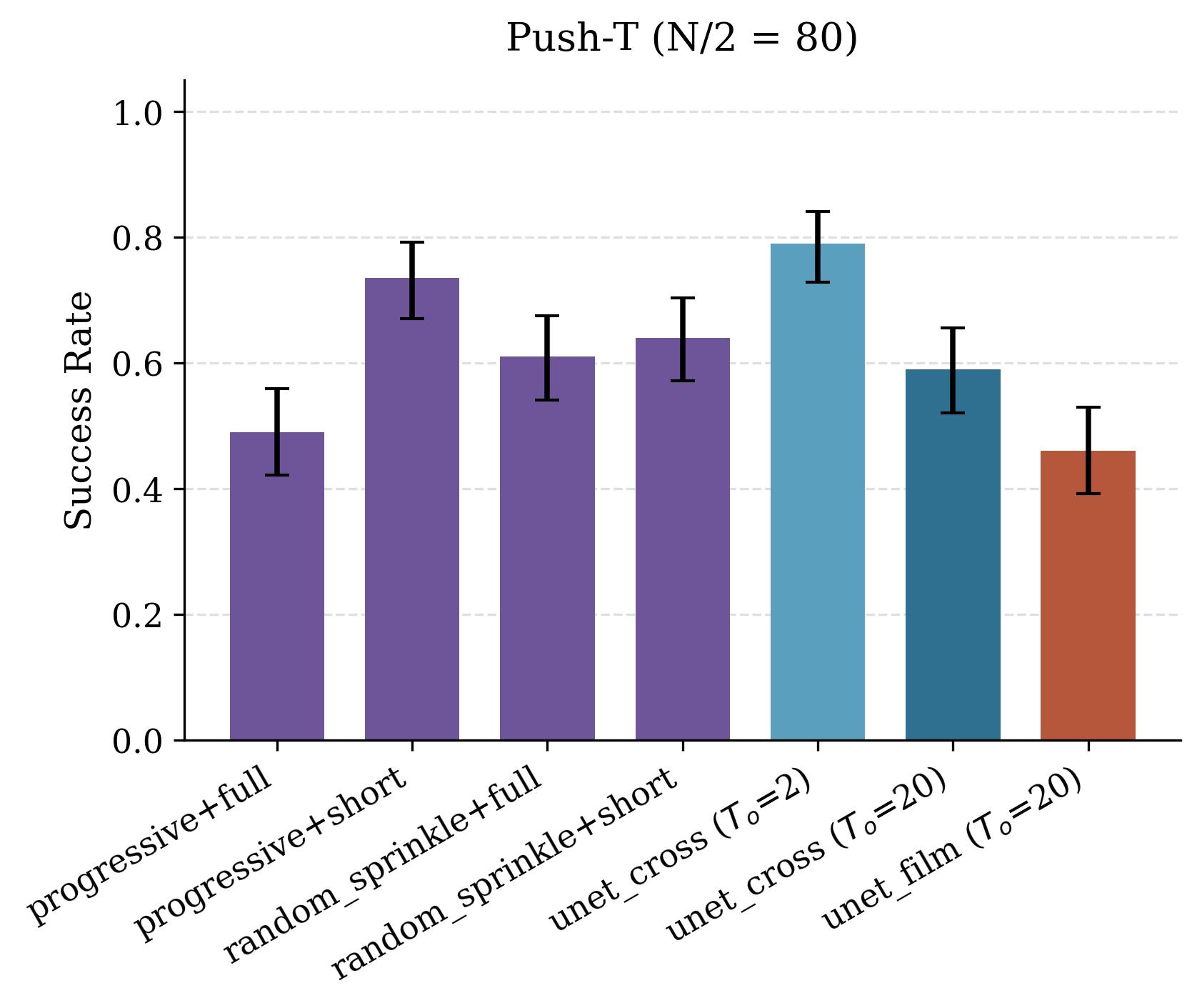}
        \caption{}
        \label{fig:push_t_variable_vs_cross_halfN}
    \end{subfigure}
    \hfill
    \begin{subfigure}[t]{0.47\textwidth}
        \centering
        \includegraphics[width=\textwidth]{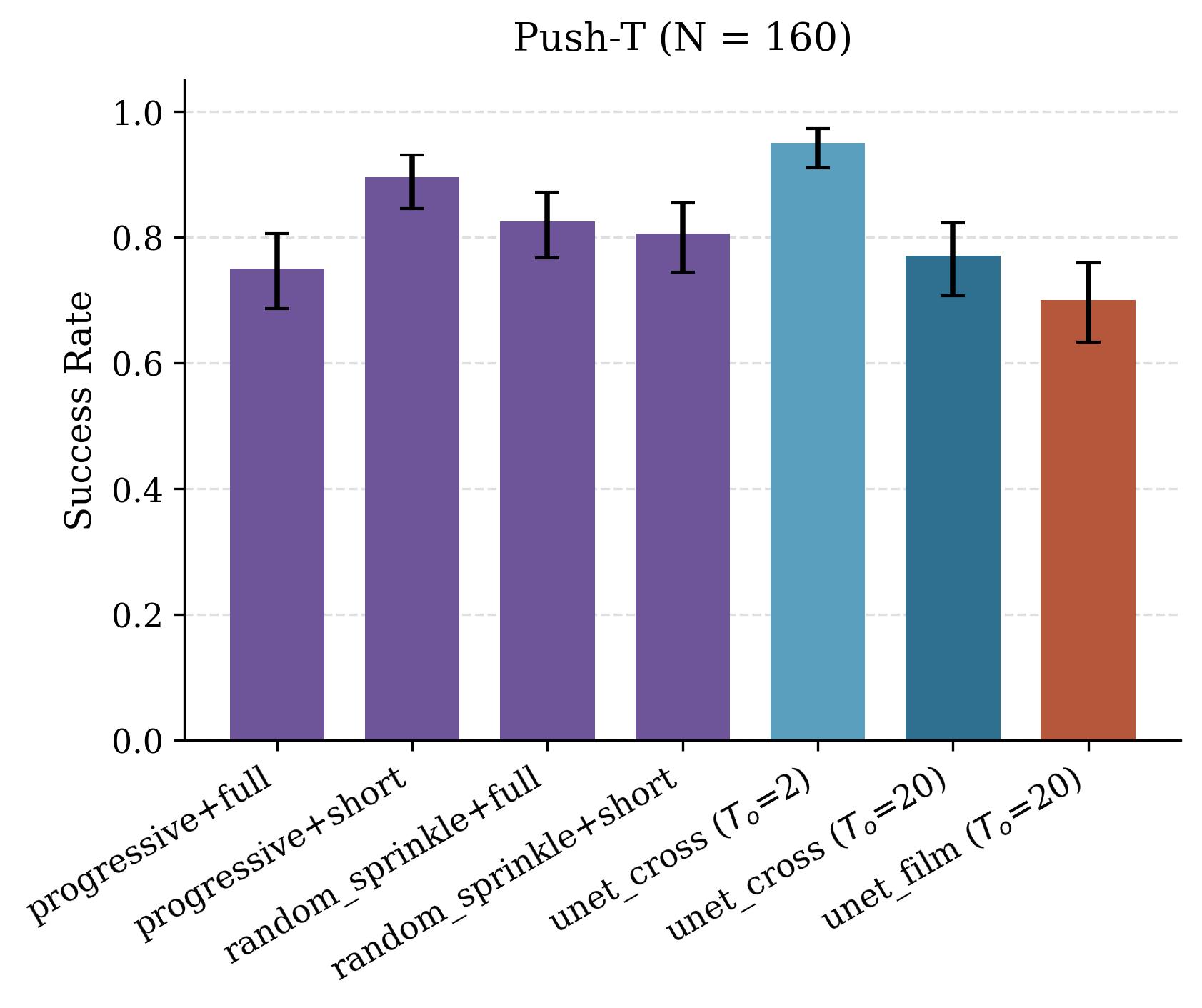}
        \caption{}
        \label{fig:push_t_variable_vs_cross_N}
    \end{subfigure}

    \medskip

    \begin{subfigure}[t]{0.47\textwidth}
        \centering
        \includegraphics[width=\textwidth]{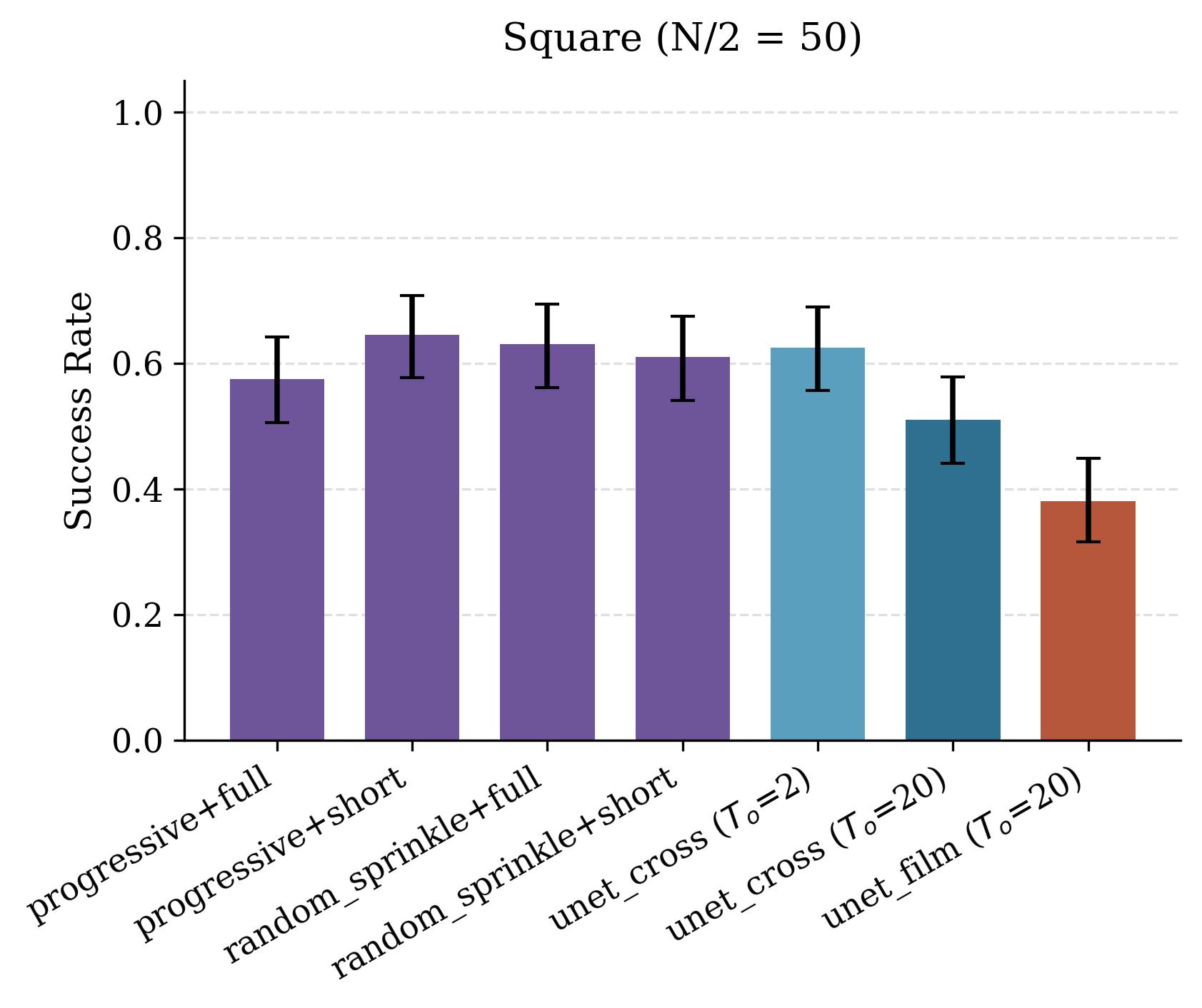}
        \caption{}
        \label{fig:square_variable_vs_cross_halfN}
    \end{subfigure}
    \hfill
    \begin{subfigure}[t]{0.47\textwidth}
        \centering
        \includegraphics[width=\textwidth]{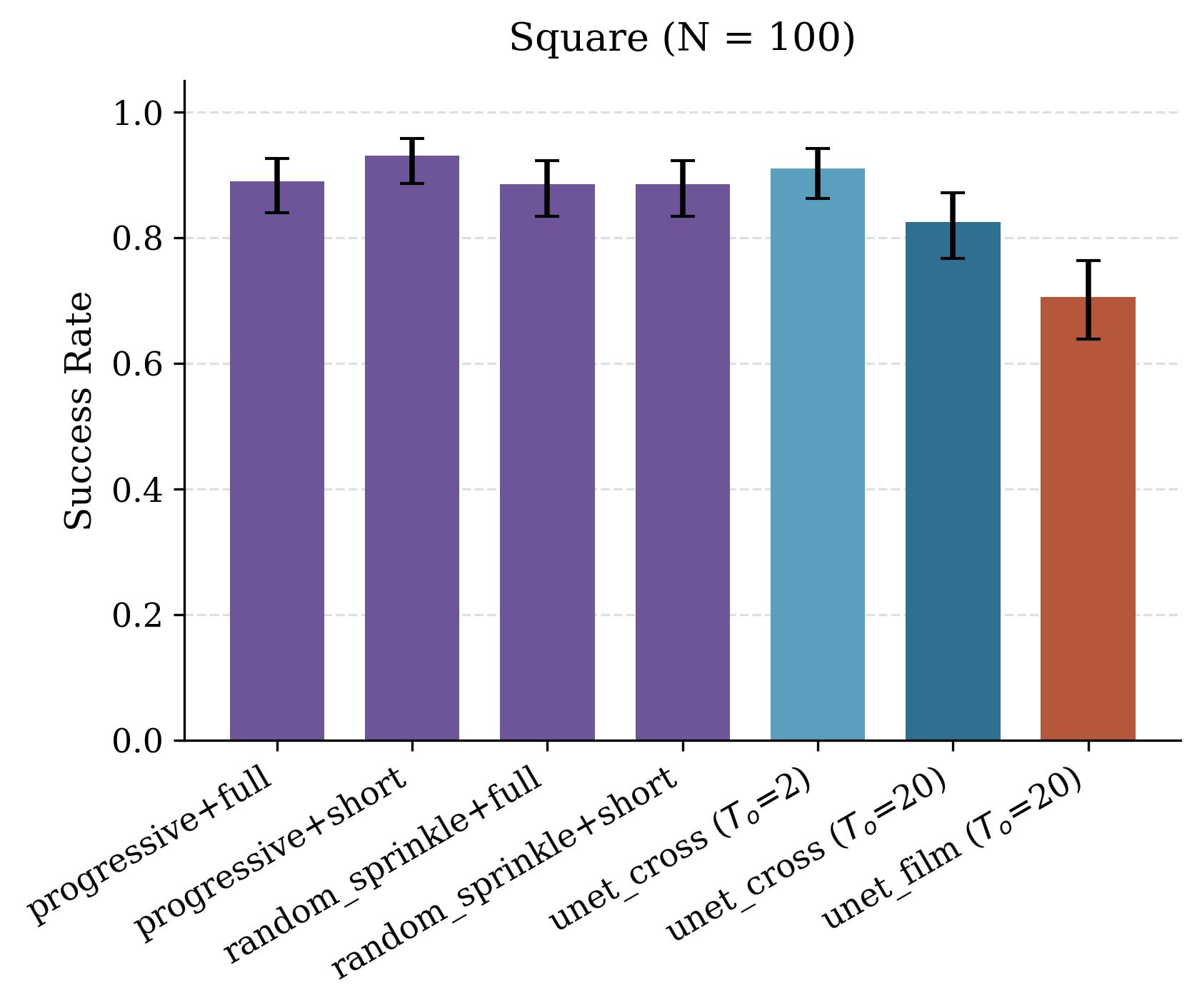}
        \caption{}
        \label{fig:square_variable_vs_cross_N}
    \end{subfigure}

    \medskip

    \begin{subfigure}[t]{0.47\textwidth}
        \centering
        \includegraphics[width=\textwidth]{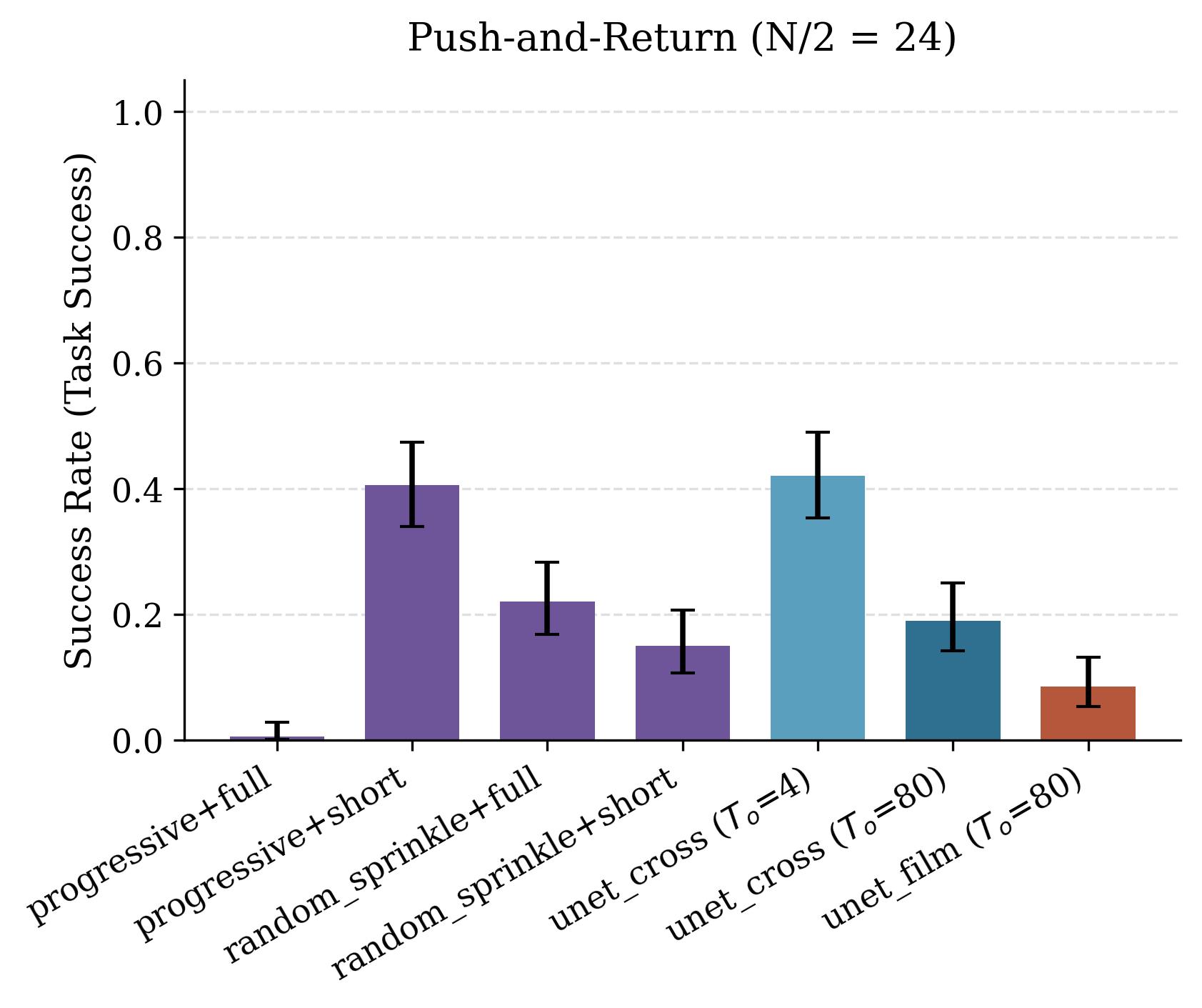}
        \caption{}
        \label{fig:push_and_return_variable_vs_cross_halfN_A}
    \end{subfigure}
    \hfill
    \begin{subfigure}[t]{0.47\textwidth}
        \centering
        \includegraphics[width=\textwidth]{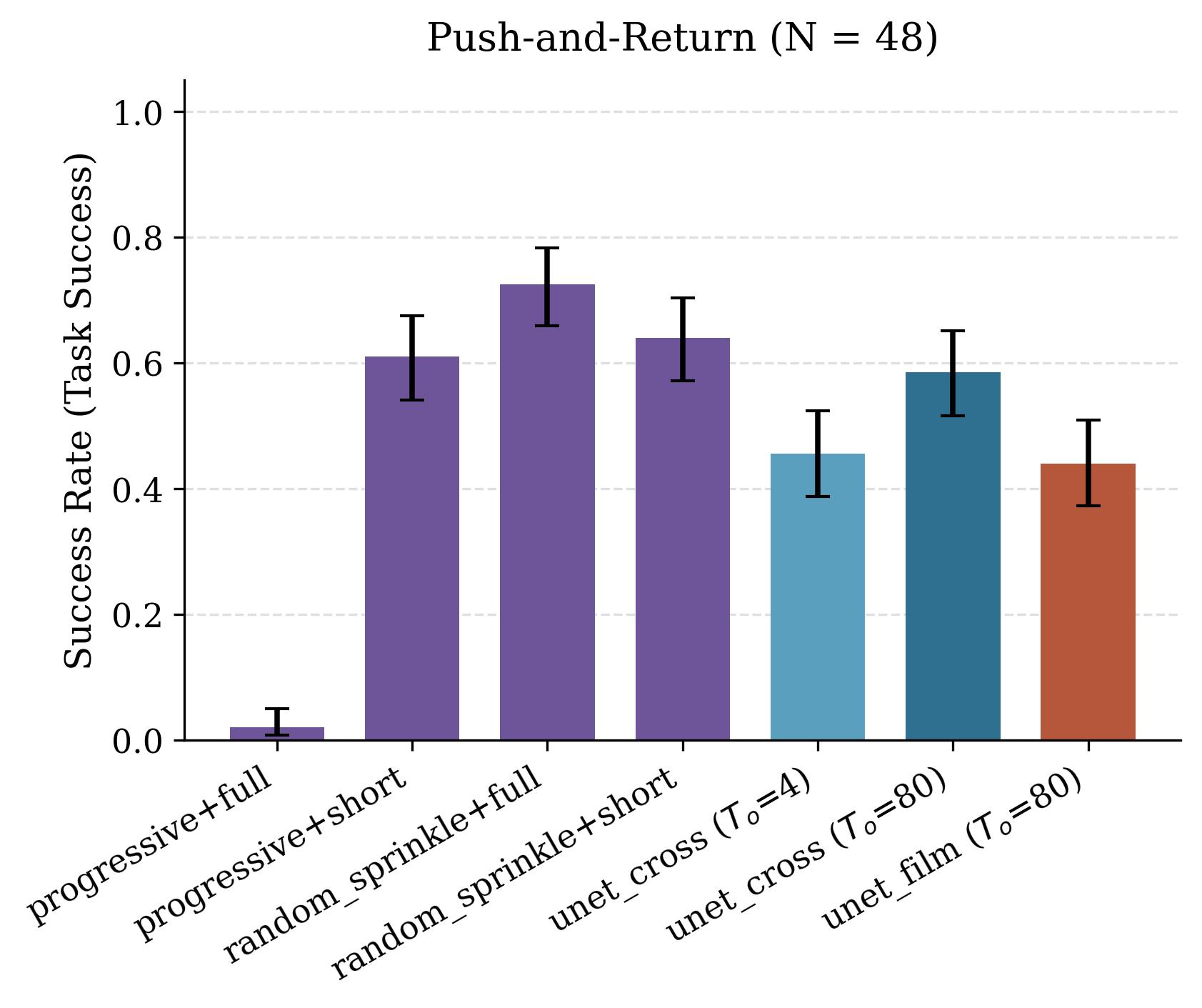}
        \caption{}
        \label{fig:push_and_return_variable_vs_cross_N_A}
    \end{subfigure}

    \caption{Comparison of \textbf{UNet+Cross-Attention}, \textbf{UNet+FiLM}, and \textbf{variable history training} method introduced above across tasks (data scale $N/2$ in left column and $N$ in right column). \textbf{Variable history} methods are trained at $T_o=20$ for \textbf{square} and \textbf{push-T} and at $T_o=80$ for \textbf{push-and-return}. For an appropriate choice of $T^{past}_p$ and $\rho_i$, the long-context model matches short-context performance in low data regime, and doesn't degrade performance in the ideal data regime.}
    \label{fig:variable_vs_cross}
\end{figure}

In Section~\ref{sec:variable_history_training}, we show significant improvements in the $N/2$ data regime. Moreover, we choose \textbf{progressive} for $\rho_i$ and \textbf{short} past action prediction horizon. In Figure~\ref{fig:variable_vs_cross}, we show results for the $N$ data regime as well as for other combinations of $\rho_i$ and $T_p^{past}$. 

We note that \textbf{progressive+short} shows performance gains in the three tasks in the $N/2$ data regime. With $N$ trajectories, it shows improvement in \textbf{push-T} and \textbf{square} (where long-context policies still show scope for improvement), and maintains performance for \textbf{push-and-return}. Note that naively scaling context length showed drop in performance in $N/2$ regime for all these tasks. However, in the $N$ data regime, we saw drop only for \textbf{push-T} and \textbf{square}, but rather rising/maintained performance for \textbf{push-and-return} as context length was increased. While \textbf{progressive+short} maintains performance in the $N$ data regime, \textbf{random sprinkle+full} shows improved performance. We thus suggest using \textbf{progressive+short} with low data, but \textbf{random sprinkle+full} when more data (sufficient for naive scaling) is available. 

\begin{figure}[!tbp]
\centering
\subfloat[]{%
    \includegraphics[width=0.48\textwidth]{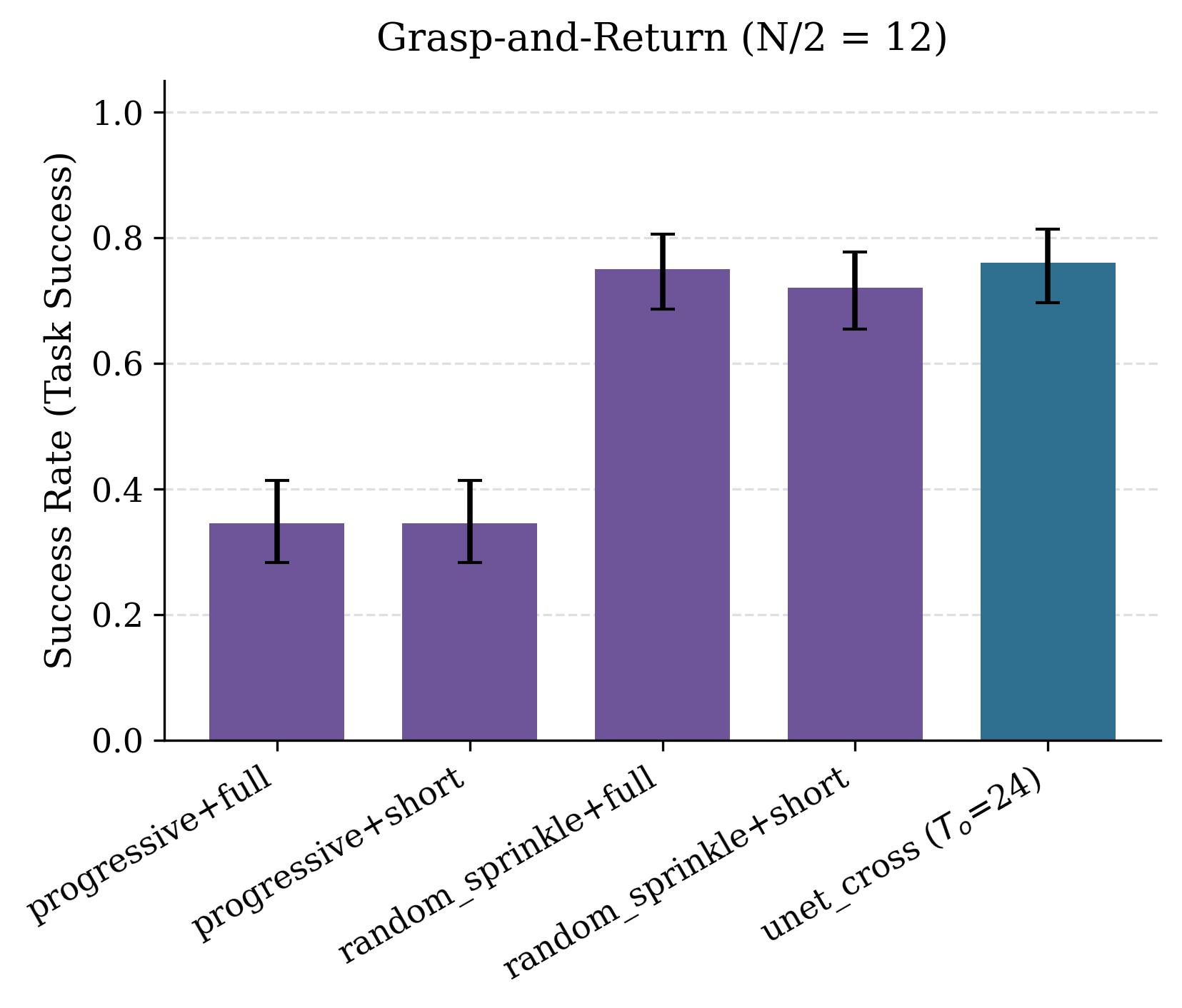}
    \label{fig:grasp_and_return_variable_vs_cross_halfN_A}
}
\hfil
\subfloat[]{%
    \includegraphics[width=0.48\textwidth]{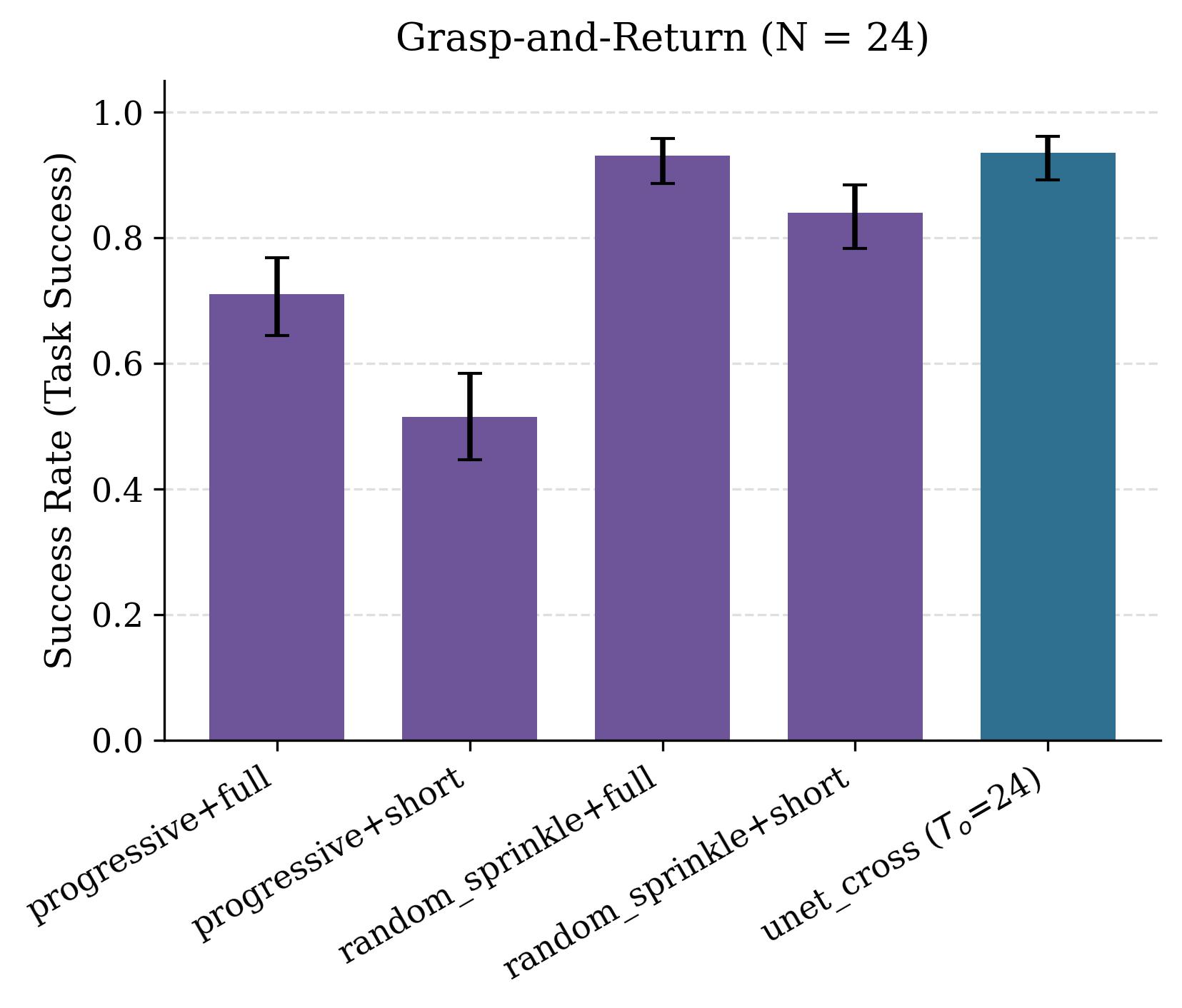}
    \label{fig:grasp_and_return_variable_vs_cross_N_A}
}
\caption{Variable history training method for \textbf{grasp-and-return} task, where naive history scaling is always sufficient for task success. Given that, we find \textbf{random sprinkle+full} to be the best method even in the $N/2$ data regime.}
\label{fig:grasp_and_return_variable_vs_cross}
\end{figure}

We confirm this by investigating these methods on the \textbf{grasp-and-return}, where, as seen in Figure~\ref{fig:grasp_and_return_cross_task_success_data_scaling}, even $N/2$ data regime shows good performance with naive context length scaling. Intuitively, \textbf{random sprinkle+full} should be the best method. That is exactly what we notice in Figure~\ref{fig:grasp_and_return_variable_vs_cross}, where in both $N/2$ and $N$ data regimes, \textbf{random sprinkle+full} performs best. 

Thus, we recommend using \textbf{progressive+short} when naive scaling performs worse compared to short context length, but \textbf{random sprinkle+full} when data is sufficient for naive history scaling. 

\subsection{Past-Token Prediction}
\label{appx:past-token_prediction}
\citet{torne2025learninglongcontextdiffusionpolicies} introduce predicting past action tokens as an auxiliary loss that helps with long-context imitation learning. Additionally, they propose freezing the observation encoder as a training time optimization technique not impacting policy performance. In this section, we independently investigate which aspect contributes to performance gains. 

We first investigate how much past-action prediction, by itself, impacts policy performance (independent of other parts of the work). Because past-action prediction wasn't proposed by \citet{torne2025learninglongcontextdiffusionpolicies} but rather present in the original work \cite{diffusion_policy}, we also investigate if predicting past actions helps with short-context policy learning. 

\begin{figure}[htbp]
    \centering
    \begin{subfigure}[t]{\linewidth}
        \centering
        \includegraphics[width=0.95\linewidth]{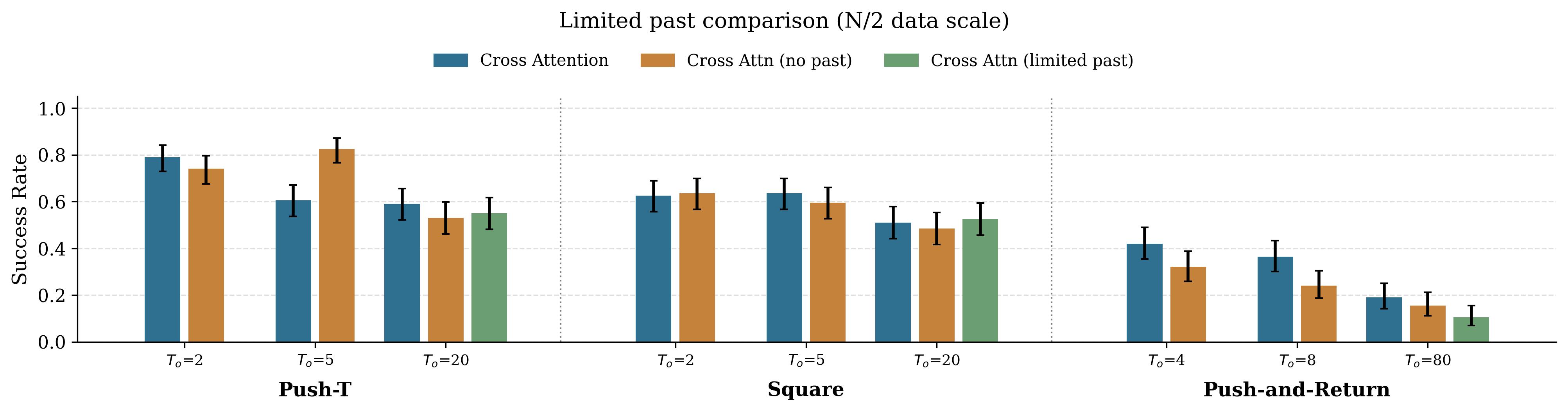}
        \caption{}
        \label{fig:limited_past_compare_halfN}
    \end{subfigure}

    \vspace{0.5em}

    \begin{subfigure}[t]{\linewidth}
        \centering
        \includegraphics[width=0.95\linewidth]{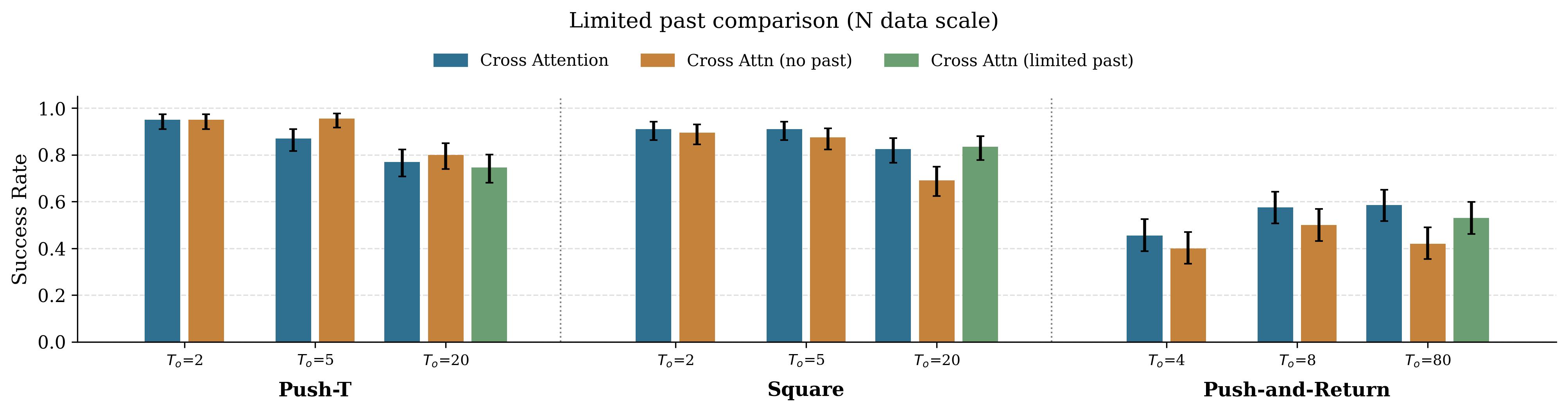}
        \caption{}
        \label{fig:limited_past_compare_N}
    \end{subfigure}
    \caption{Results of a study investigating how much past action prediction, a design choice introduced by \citet{diffusion_policy}, helps policy learning with the \textbf{UNet+Cross-Attention} architecture. While advantages at long context lengths are evident for \textbf{square} and \textbf{push-and-return} in the $N$ data regime, performance is more similar in the limited data case. Thus, overall patterns are unclear.}
    \label{fig:past_action_prediction}
\end{figure}

Investigating Figure~\ref{fig:past_action_prediction}, we find that by itself, predicting past actions presents limited advantages. While advantages at long context lengths are evident for \textbf{square} and \textbf{push-and-return} in the $N$ data regime, performance is more similar in the limited data case. There are also no noticeable trends for short-context policy performance. Given the limited advantage which is not seen across data scale, we conclude that the benefit of predicting past actions by itself is unclear for long-context learning. 

\begin{figure}[!tbp]
    \centering
    \includegraphics[width=0.95\linewidth]{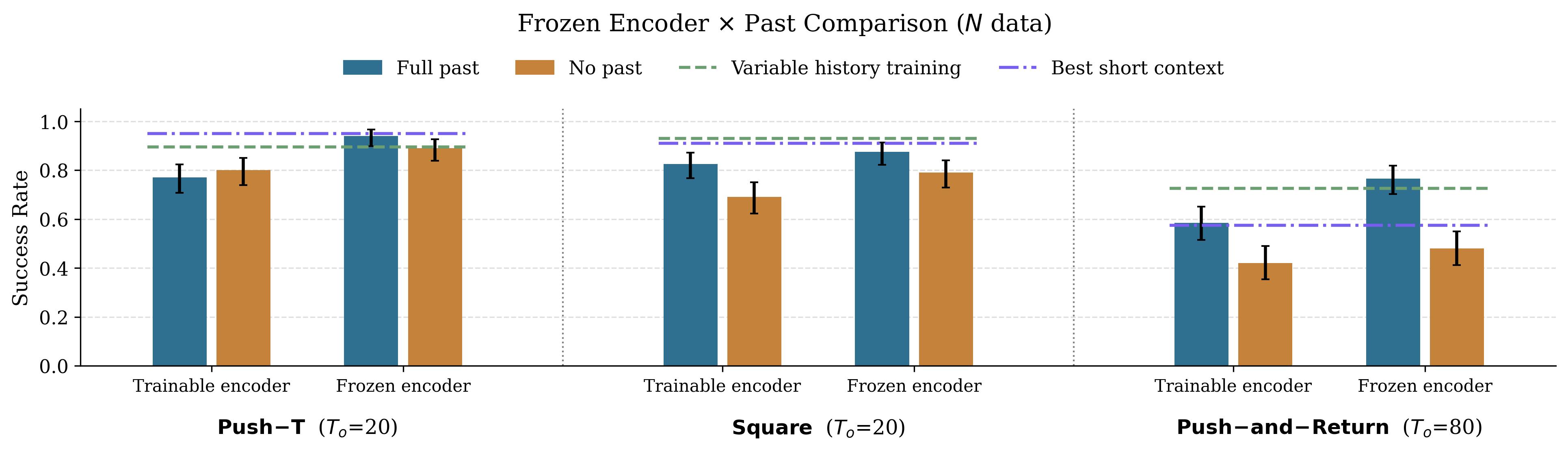}
    \caption{Results comparing trainable and observation encoder frozen from short context lengths with past action prediction. Past-action prediction and frozen observation encoder seem to be driving each others' success, but \textbf{push-and-return} is an exception. \vspace{-\baselineskip}}
    \label{fig:frozen_encoder_past_comparison_N}
\end{figure}

Next, we continue the discussion from Section~\ref{subsec:revisiting_ptp}, where we investigate the relative difference in policy learning between trainable and frozen encoder. While Figure~\ref{fig:frozen_encoder_past_comparison_N_half} shows the comparison in the $N/2$ data regime, we now show comparison in the $N$ data regime in Figure~\ref{fig:frozen_encoder_past_comparison_N}. We find that at both data scales, past-action prediction and freezing the observation encoder seem to be driving each others' success - freezing the observation encoder is not just a training technique, but seems to also be responsible for improving policy performance at least to some extent. The variable history comparison presented in Figures~\ref{fig:frozen_encoder_past_comparison_N_half} and \ref{fig:frozen_encoder_past_comparison_N} picks the value with the best hyperparameter (between choice of \textbf{random sprinkle} v/s \textbf{progressive}).  

As a final note, we state that \citet{torne2025learninglongcontextdiffusionpolicies} also propose an inference time verification technique, where multiple action chunks are sampled from the policy, and the chunk with past actions most consistent with the actually executed past actions is executed. We exclude this technique from our analysis because it reduces inference speed as we evaluate a large number of policies, and by authors' own admission, this helps by upto only 5\%. We did run some comparisons presented in Table~\ref{tab:verification_results} and found that success improvement from this technique was statistically insignificant, and thus excluded it from our comparisons. 

\begin{table}[h]
\centering
\caption{Performance with and without PTP inference time verification. Policy trained in $N$ data regime at $T_o = 20$ with frozen observation encoder from the best short-context policy for that task.}
\begin{tabular}{lcc}
\toprule
Task & With Verification & W/O Verification \\
\midrule
Square & 176/200 & 175/200 \\
Push-T & 190/200 & 188/200 \\
\bottomrule
\end{tabular}
\label{tab:verification_results}
\end{table}

\FloatBarrier
\subsection{Hardware Validation}
\label{appx:hardware_experiments}
Consider the \textbf{marshmallows} task described in Section~\ref{appx:task_details}. We naively scale context length to $T_o=92$ for the 3 architectures we study in Section~\ref{sec:cross_attention_conditioning}. Results are shown in Figure~\ref{fig:hardware_marshmallows}. 

This task has been adapted from \citet{mark2026bpplongcontextrobotimitation}. While they use 250 demonstrations and show that their VLM based filtering trick is necessary for success, we are able to achieve high success rates with simple naive scaling and just 100 demonstrations. While our exact setups are different due to different robot embodiments and that could certainly explain any differences in our outcomes, we believe that grasping from a bowl full of marshmallows allows room for manipulation error while still succeeding (as off grasps are also likely to capture an object). 

Figure~\ref{fig:hardware_marshmallows} shows that all our conditioning architectures perform similarly. We speculate this could be because 100 trajectories are more than enough for a task like this. However, we do observe qualitative differences. We find \textbf{UNet+Cross-Attention} to be smoother, clearing bowl boundaries with larger margins, and exhibiting more robust re-grasps when needed. \textbf{DiT}, on the other hand, frequently re-grasps even when not necessary, and the gripper bumps into boundary of the bowl/doesn't clear it with high margins. The movement is also not as smooth. \textbf{UNet+FiLM} qualitatively sits between the other two choices. Quantitative evaluation of such metrics in hardware remains an open challenge for robotics.

\end{document}